\documentclass[10pt,twocolumn,letterpaper]{article}

\usepackage[pagenumbers]{iccv} 

\usepackage[accsupp]{axessibility}  

\usepackage{multirow}

\usepackage[ruled]{algorithm2e} 
\usepackage{algpseudocode}

\usepackage{kotex}
\newcommand{\sunghyun}[1]{{\textcolor[rgb]{0.6,0.0,0.6}{[sunghyun: #1]}}}

\renewcommand{\paragraph}[1]{{\vspace{2pt}\noindent\textbf{#1}}}

\definecolor{iccvblue}{rgb}{0.21,0.49,0.74}
\usepackage[pagebackref,breaklinks,colorlinks,allcolors=iccvblue]{hyperref}

\usepackage{color, colortbl}
\usepackage{float}

\definecolor{cBlue}{HTML}{1852CC}
\definecolor{cBlue2}{HTML}{3fb8b8}
\definecolor{cBlue3}{HTML}{02ffff}
\definecolor{cBlue4}{HTML}{00b5ff}
\definecolor{cRed}{HTML}{D62728}
\definecolor{cRed2}{HTML}{ED521F}
\definecolor{cRed3}{HTML}{F69C40}
\definecolor{cRed4}{HTML}{fa4d3a}
\definecolor{cGreen}{HTML}{2CA02C}
\definecolor{cGreen2}{HTML}{3fdf3f}
\definecolor{cPink}{HTML}{ED1FD2}
\definecolor{cWhite}{HTML}{ffffff}
\definecolor{Violet}{HTML}{b05cff}
\definecolor{Gray}{gray}{0.9}
\definecolor{Salmon}{HTML}{FF7E79}
\definecolor{Orchid}{HTML}{7A81FF}
\definecolor{cBlue5}{HTML}{409BA0}
\definecolor{cPink2}{HTML}{CB2CED}
\definecolor{cPink3}{HTML}{ED1D81}
\definecolor{PastelPink}{HTML}{FC94AF}

\newcommand{\js}[1]{{\color{cGreen}{#1}}}

\def\paperID{15419} 
\def\confName{ICCV}
\def\confYear{2025}

\title{Addressing Text Embedding Leakage in Diffusion-based Image Editing}

\author{
$\text{Sunung Mun}^{1 *}$ \quad\quad 
$\text{Jinhwan Nam}^{1 *}$ \quad\quad
$\text{Sunghyun Cho}^{1,2}$ \quad\quad
$\text{Jungseul Ok}^{1,2{\dagger}}$\\
\\
$\text{Graduate School of AI, POSTECH}^1$, \quad\quad
$\text{Dept. of CSE, POSTECH}^2$\\
{\tt\small \{mtablo, njh18, s.cho, jungseul\}@postech.ac.kr}
}

\newcommand{\benchmark}{ALE-Bench}

\begin{document}

\twocolumn[{
\renewcommand\twocolumn[1][]{#1}
\maketitle

\begin{center}
    \centering
    \vspace{-0.7cm}
    \captionsetup{type=figure}
    \scalebox{1}{\includegraphics[width=0.98\textwidth]{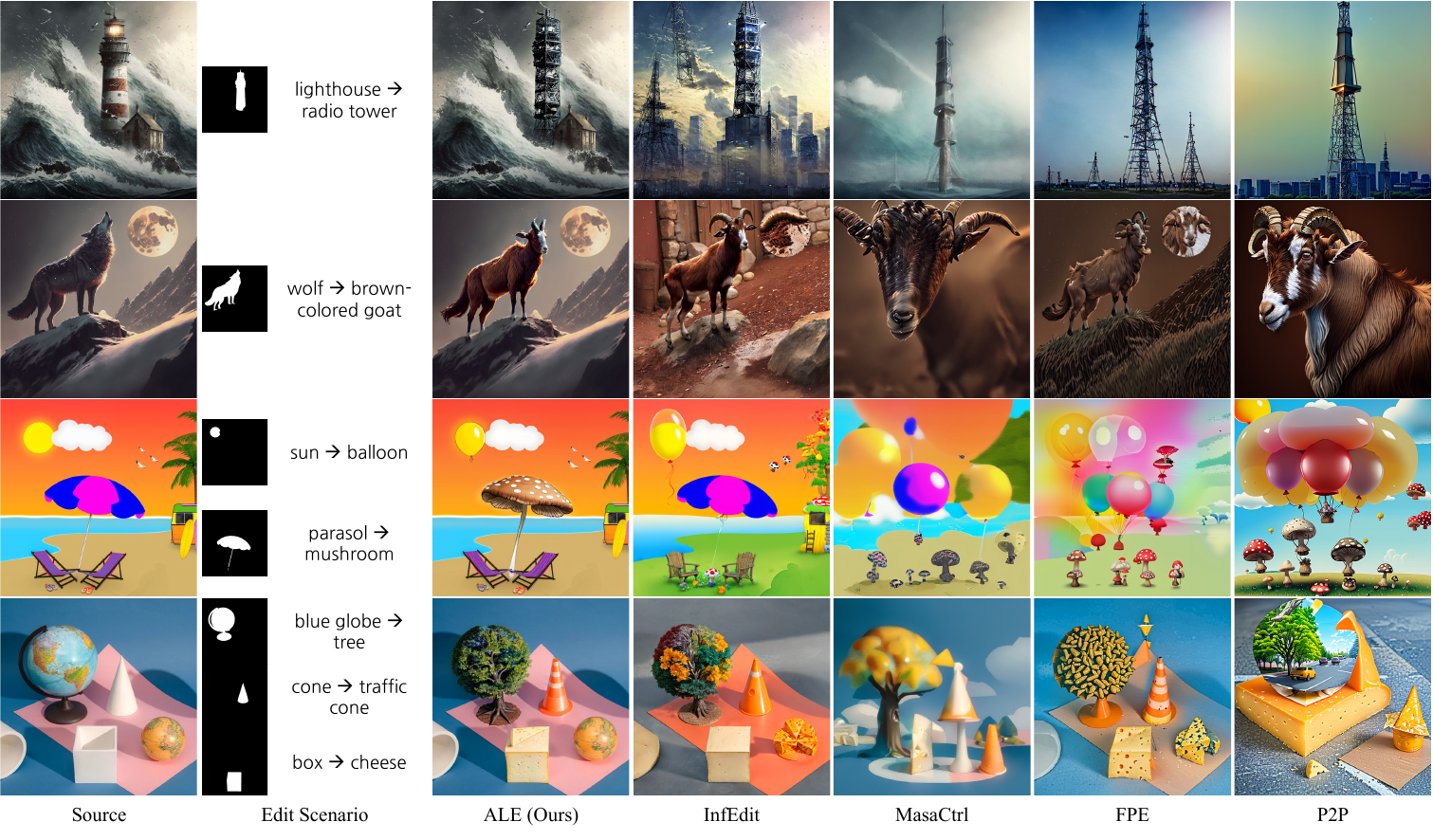}}
    \vspace{-0.3cm}
    \captionof{figure}{
Qualitative comparison of tuning-free image-editing methods.
Left to right: (1) source image, (2) visualization-only binary masks of the source objects and corresponding edit prompts (source $\xrightarrow{}$ target), and (3–7) results from our ALE and four baselines.  
The masks are not provided to any method; they are shown solely to indicate which region is supposed to change.  
Rows 1–2 illustrate single-object edits where baselines spill changes outside the intended area or distort geometry, while ALE keeps the background intact.  
Rows 3–4 demonstrate multi-object edits: baselines often entangle attributes, whereas ALE preserves each attribute in its designated region, yielding leakage-free results.
    }
        \label{fig:main}
\end{center}
}]

\let\thefootnote\relax\footnotetext{
    $^*$ Equal contribution; {$^\dagger$} Correspondence to 
    \href{mailto:jungseul@postech.ac.kr}{jungseul@postech.ac.kr}
}

\let\thefootnote\relax\footnotetext{
    Project page: 
    \href{https://mtablo.github.io/ALE_Edit_page/}{https://mtablo.github.io/ALE\_Edit\_page/}
}

\begin{abstract}

Text-based image editing, powered by generative diffusion models, 
lets users modify images through natural-language prompts and has dramatically simplified traditional workflows.
Despite these advances, current methods still suffer from a critical problem: \emph{attribute leakage}, where edits meant for specific objects unintentionally affect unrelated regions or other target objects. 
Our analysis reveals the root cause as the semantic entanglement inherent in End-of-Sequence (EOS) embeddings generated by autoregressive text encoders, which indiscriminately aggregate attributes across prompts.
To address this issue, we introduce \emph{Attribute-Leakage-free Editing} ({ALE}), a framework that tackles attribute leakage at its source. 
ALE combines \emph{Object-Restricted Embeddings} (ORE) to disentangle text embeddings, \emph{Region-Guided Blending for Cross-Attention Masking} (RGB-CAM) for spatially precise attention, and \emph{Background Blending} (BB) to preserve non-edited content. 
To quantitatively evaluate attribute leakage across various editing methods, we propose the \emph{Attribute-Leakage Evaluation Benchmark} (\benchmark{}), featuring comprehensive editing scenarios and new metrics. 
Extensive experiments show that ALE reduces attribute leakage by large margins, thereby enabling accurate, multi-object, text-driven image editing while faithfully preserving non-target content.
\end{abstract}

\section{Introduction}
\label{sec:intro}

Text-based image editing, where users modify existing images via natural language prompts, has emerged as a powerful alternative to traditional manual editing. Conventional editing workflows typically demand significant manual effort and domain expertise~\cite{sangkloy2017scribbler, zhu2016generative, isola2017image, brock2018large}. In contrast, recent advancements leveraging generative diffusion models have substantially simplified the editing process~\cite{meng2021sdedit, elarabawy2022direct, huang2024paralleledits, guo2024focus}. These models enable users to perform high-quality edits through intuitive textual prompts, making image editing more accessible and flexible.

Despite these advances, existing text-based editing methods frequently suffer from a critical limitation: \emph{attribute leakage}, where edits intended for specific objects inadvertently affect unrelated regions within the image. 
Attribute leakage can be categorized into two distinct types: \emph{Target-External Leakage (TEL)}, where attributes of a target object unintentionally affect non-target regions, and \emph{Target-Internal Leakage (TIL)}, where attributes intended for one target object inadvertently influence another target object within the same editing prompt. 
To mitigate this, recent studies attempt to spatially constrain editing effects by manipulating cross-attention maps~\cite{cao2023masactrl, xu2023inversion}. 
However, as illustrated in Figure~\ref{fig:main}, even state-of-the-art methods~\cite{cao2023masactrl, xu2023inversion, zou2024towards, hertz2022prompt} continue to exhibit significant TEL in single-object editing scenarios, and both TEL and substantial TIL in more complex, \emph{multi-object} editing scenarios.

Our analysis highlights that attribute leakage fundamentally stems from overlooked issues of the entanglement of text embeddings, specifically those associated with the \emph{End-of-Sequence (EOS)} tokens, building upon recent study~\cite{hu2024token}.
Most text-based editing pipelines employ autoregressive text encoders such as CLIP~\citep{radford2021learning}, which append EOS tokens to text prompts until reaching a fixed length (e.g., 77 tokens) to generate embeddings.
Consequently, EOS embeddings inherently aggregate information from \emph{all tokens} within the prompt and attend indiscriminately across image regions via cross-attention layers, exacerbating both TEL and TIL.
While recent studies~\cite{hu2024token} attempt to alleviate leakage stemming from EOS embedding entanglement, our analysis demonstrates that these approaches are insufficient to fully eliminate attribute leakage (see Section~\ref{sec:problem_analysis}).

To address these limitations, we propose a novel framework, \emph{Attribute-Leakage-Free Editing (ALE)}, primarily consisting of three complementary components: Object-Restricted Embeddings (ORE), Region-Guided Blending for Cross-Attention Masking (RGB-CAM), and Background Blending (BB).
ORE assigns distinct, semantically isolated embeddings to each object in the prompt, explicitly avoiding embedding entanglement. 
RGB-CAM enhances spatial precision in cross-attention maps by leveraging segmentation masks, restricting attention solely to intended regions.
BB preserves the structural integrity of non-edited regions by integrating latents from the source image.

Furthermore, to systematically quantify attribute leakage, we introduce the \emph{Attribute-Leakage Evaluation Benchmark (\benchmark{})}, a specialized benchmark designed explicitly for leakage evaluation in multi-object editing. Existing benchmarks predominantly focus on single-object scenarios, lacking metrics for evaluating attribute leakage comprehensively~\cite{wang2023imagen, Zhang2023MagicBrush, ju2023direct}. 
\benchmark{} covers a diverse range of editing scenarios, including multi-object editing in various edit types. 
We also propose two novel evaluation metrics: \emph{Target-External Leakage Score (TELS)} and \emph{Target-Internal Leakage Score (TILS)}, explicitly quantifying TEL and TIL, respectively.

In summary, our contributions are: 
\begin{enumerate} 
    \item Identifying the previously overlooked role of EOS embeddings as a fundamental cause of attribute leakage in text-based image editing (Section~\ref{sec:problem_analysis}). 
    \item Proposing the novel \emph{Attribute-Leakage-Free Editing (ALE)} framework, which specifically addresses leakage induced by EOS embeddings (Section~\ref{sec:method}). 
    \item Introducing a comprehensive benchmark, \emph{\benchmark{}}, along with novel metrics (\emph{TELS}, \emph{TILS}) designed explicitly to quantify attribute leakage in multi-object editing scenarios (Section~\ref{sec:exp}). 
\end{enumerate}

\section{Attribute Leakage Problem and Analysis}
\label{sec:problem_analysis}

In this section, we discuss the \emph{attribute leakage} problem in multi-object text-based image editing and highlight limitations of existing methods.
Specifically, Section~\ref{sec:image editing} formally introduces multi-object text-based image editing and briefly describes dual-branch frameworks commonly used for such tasks.
Section~\ref{sec:attribute leakage} defines attribute leakage and categorizes it into two types: TEL and TIL.
Finally, Section~\ref{sec:causes_of_leakage} analyzes embedding entanglement, particularly from EOS tokens, as the primary cause of attribute leakage and explains why existing methods fail to adequately address this issue.

\subsection{Multi-Object Text-based Image Editing}
\label{sec:image editing}

\begin{figure}[ht!]
    \centering
    \begin{subfigure}[b]{0.32\linewidth}
        \includegraphics[width=\linewidth]{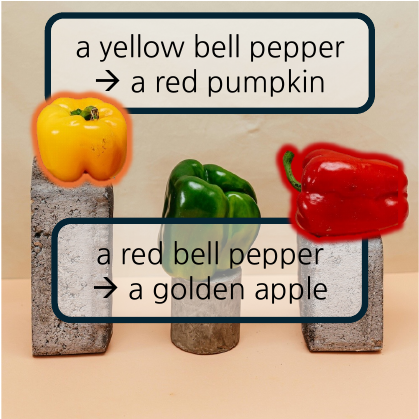}
        \caption{Source Image}
        \label{fig:sub:source_edit}
    \end{subfigure}
    \begin{subfigure}[b]{0.32\linewidth}
        \includegraphics[width=\linewidth]{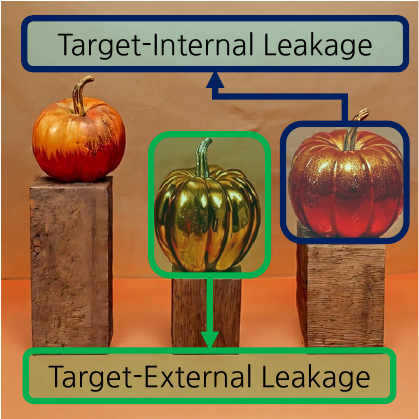}
        \caption{Leakage Example}
        \label{fig:sub:leakage}
    \end{subfigure}
    \begin{subfigure}[b]{0.32\linewidth}
        \includegraphics[width=\linewidth]{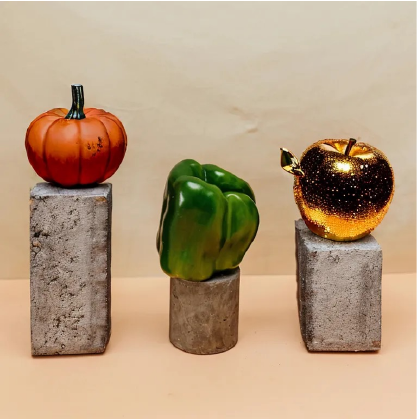}
        \caption{Reference Result}
        \label{fig:sub:reference}
    \end{subfigure}
    \caption{
    Illustration of attribute leakage in image editing.
    (a) The source image and editing prompts.
    (b) An editing result exhibiting attribute leakage. 
    Regions highlighted in green indicate \emph{target-external leakage}, where the editing spills into non-target (e.g., the green bell pepper turns into a golden pumpkin). 
    Regions in blue show \emph{target-internal leakage}, where the attributes of ``a red pumpkin'' undesirably influence the appearance of the ``golden apple'' region.
    (c) A reference image showing the desired editing result without attribute leakage.
    }
    \label{fig:leakage_example}
    \vspace{-0.5cm}
\end{figure}

\emph{Text-based image editing} modifies specific regions of a source image according to textual prompts describing desired changes. 
Formally, given a source image $x^{\text{src}}$ and a textual prompt pair $(y^{\text{src}},y^{\text{tgt}})$—where $y^{\text{src}}$ specifies the source objects to be edited (i.e., objects to be replaced or modified), and $y^{\text{tgt}}$ specifies the target objects (i.e., new or modified objects to appear)—the goal is to generate an edited image $x^{\text{tgt}}$.
Ideally, $x^{\text{tgt}}$ reflects modifications described by $y^{\text{tgt}}$ exclusively within regions indicated by $y^{\text{src}}$, leaving all other areas unchanged. 
When edits involve multiple objects simultaneously, the task is classified as \emph{multi-object editing}.

Practically, multi-object editing prompts can be decomposed into individual object-level prompt pairs $[(y^{\text{src}}_i,y^{\text{tgt}}_i)]_{i=1}^K$, typically via language models or noun-chunk parsers~\cite{brown2020language, grattafiori2024llama}. 
Our research specifically focuses on scenarios involving up to $K=3$ objects. 
For example, consider the editing scenario in Figure~\ref{fig:sub:source_edit}. 
Given the prompt pair $(y^{\text{src}},y^{\text{tgt}})=$ (``a yellow bell pepper and a red bell pepper'', ``a red pumpkin and a golden apple''), it can be decomposed into object-level prompts $[(y_{i}^{\text{src}}, y_i^{\text{tgt}})]_{i=1}^2=$ [(``a yellow bell pepper'', ``a red pumpkin''), ``a red bell pepper'', ``a golden apple'')]. 
The resulting image $x^{\text{tgt}}$ should contain a red pumpkin and a golden apple, precisely aligned to corresponding regions.

To perform such precise edits, diffusion-based editing methods encode textual prompts into embeddings that guide the editing process through cross-attention layers. These layers spatially align text-described attributes with corresponding regions in the image. Achieving accurate alignment, however, requires effectively preserving the original spatial structure while synthesizing novel visual attributes. To this end, recent methods utilize \emph{dual-branch editing frameworks}, which have become popular due to their ability to simultaneously retain the original image layout and introduce new content~\cite{cao2023masactrl, xu2023inversion, liu2024towards}.

Dual-branch frameworks operate through two parallel pathways—a \emph{source branch} and a \emph{target branch}—using the same pretrained text-to-image diffusion model.
Specifically, the \emph{source branch} reconstructs the original image $x^{\text{src}}$ guided by the source textual prompt $y^{\text{src}}$, thereby capturing structural and spatial information inherent in $x^{\text{src}}$.
The \emph{target branch}, in parallel, synthesizes new visual attributes guided by the target textual prompt $y^{\text{tgt}}$.
Structural consistency is maintained by injecting intermediate self-attention layer components—such as queries and keys—computed from the source branch into the corresponding self-attention layers of the target branch. 
Despite their strengths, dual-branch frameworks face fundamental challenges, particularly \emph{attribute leakage}, caused by embedding entanglement involving EOS tokens, as analyzed in subsequent sections.


\subsection{Attribute Leakage}
\label{sec:attribute leakage}

\emph{Attribute leakage} is a critical challenge in multi-object text-based image editing, characterized by unintended propagation of attributes from target objects to unrelated regions or other target objects.
Formally, given a source image $x^\text{src}$ and object-level prompt pairs $[(y_i^\text{src}, y_i^\text{tgt})]_{i=1}^{K}$, attribute leakage occurs when modifying an object from its original description $y_i^\text{src}$ to a new target description $y_i^\text{tgt}$ unintentionally impacts regions or objects not specified by the editing prompt.
Attribute leakage can be categorized into two distinct types:
\begin{itemize}
    \item \emph{Target-External Leakage (TEL)}: This occurs when editing an object specified by $(y_i^\text{src}, y_i^\text{tgt})$ unintentionally affects regions not described by the prompt pair (i.e., non-target regions).
    For example, as highlighted by the green region in Figure~\ref{fig:sub:leakage}, editing $y_2^\text{src}=\text{``a red bell pepper''}$ to $y_2^\text{tgt}=\text{``a golden apple''}$ inadvertently transforms an unrelated ``green bell pepper'' into a golden object.
    
    \item \emph{Target-Internal Leakage (TIL)}: This occurs when editing an object specified by $(y_i^\text{src}, y_i^\text{tgt})$ unintentionally affects another target object specified by a different prompt pair $(y_j^\text{src}, y_j^\text{tgt})$, where $i \neq j$.
    For instance, as illustrated by the blue region in Figure~\ref{fig:sub:leakage}, editing $y_1^\text{src}=\text{``a yellow bell pepper''}$ to $y_1^\text{tgt}=\text{``a red pumpkin''}$ inadvertently impacts another target object described by $y_2^\text{tgt}=\text{``a golden apple''}$, causing it to appear as a mixture of red and golden pumpkin-like attributes.
\end{itemize}
Effectively mitigating both TEL and TIL is essential to achieving precise, user-intended edits, as exemplified by the desired reference result in Figure~\ref{fig:sub:reference}.

\subsection{Causes of Attribute Leakage}
\label{sec:causes_of_leakage}
\begin{figure}[t!]
    \centering
    
    \begin{subfigure}{0.97\linewidth}
        \includegraphics[width=\linewidth]{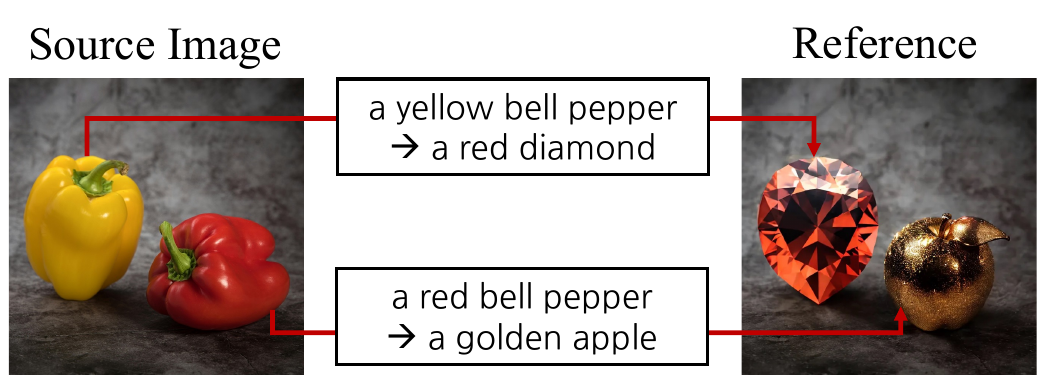}
        \caption{Source image, edit prompt, and reference result image}
    \end{subfigure}
    
    \par\medskip
    \begin{subfigure}{0.97\linewidth}
        \includegraphics[width=\linewidth]{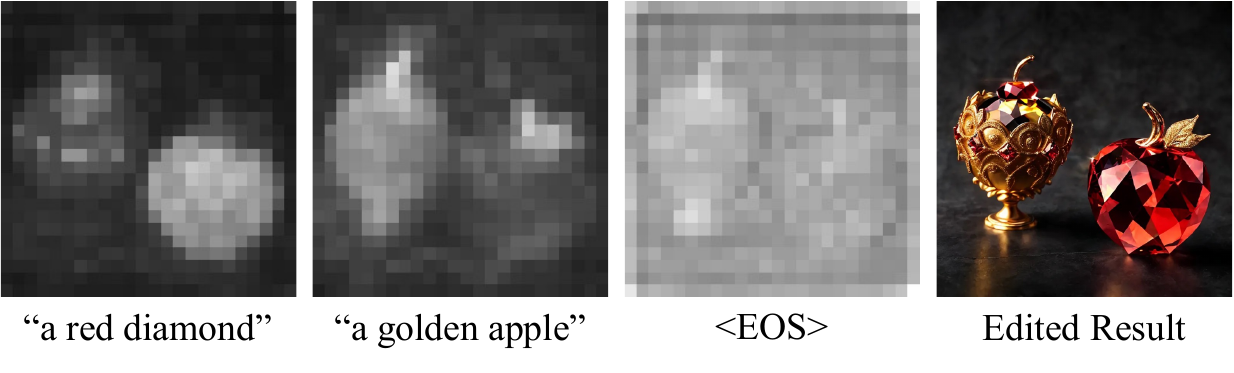}
        \caption{Cross-attention maps and an edited result of ETS}
        \label{fig:sub:ets}
    \end{subfigure}

    \par\medskip
    \begin{subfigure}{0.97\linewidth}
        \includegraphics[width=\linewidth]{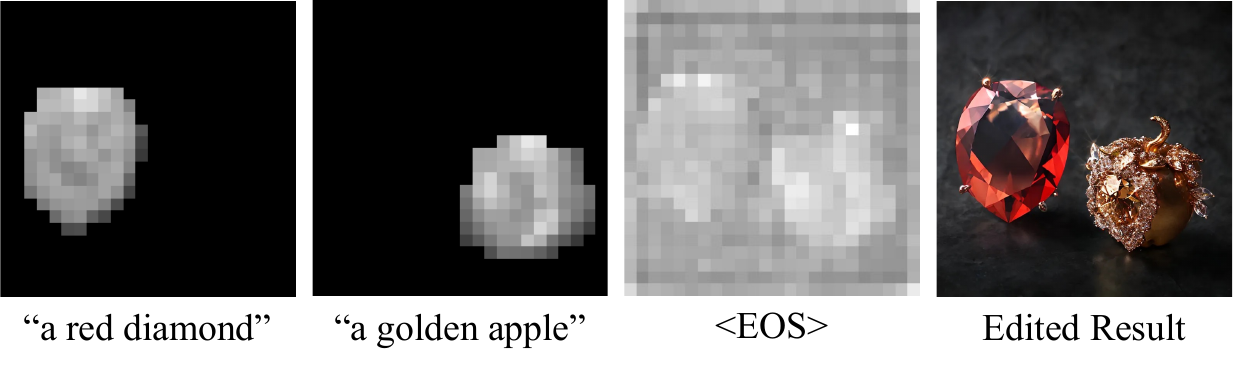}
        \caption{Cross-attention maps and an edited result of ETS + Masking}
        \label{fig:sub:ets_masking}
    \end{subfigure}

    \caption{
    Illustration of attribute leakage from EOS embeddings and misaligned cross-attention.
    (b) ETS with object-wise embeddings shows spatial misalignment: embeddings of one object (e.g., ``a red diamond'') can influence unrelated regions.
    (c) Adding cross-attention masking correctly localizes object-wise embeddings but fails to suppress leakage from EOS embeddings (e.g., diamond-like decoration on the apple).
    Cross-attention maps are averaged across timesteps and summed over tokens (or padded EOS tokens).
    }
    \label{fig:limitations_existing}
    \vspace{-0.5cm}
\end{figure}

Attribute leakage primarily arises from \emph{embedding entanglement}, which occurs during prompt encoding.
Widely used text encoders, such as CLIP, encode tokens autoregressively, causing embeddings of later tokens to unintentionally accumulate mixed semantics from preceding attributes or objects.
To mitigate this entanglement, some approaches introduce object-wise embeddings by parsing prompts into distinct noun-phrase spans and encoding each segment independently~\cite{feng2022training}.
However, these methods only address entanglement among original tokens in the prompt, failing to adequately resolve entanglement involving EOS tokens.

Since CLIP pads prompts to a fixed length using EOS tokens, the EOS embeddings inevitably aggregate semantic information from multiple attributes and objects.
For instance, when encoding the prompt ``a red diamond and a golden apple'', the EOS embeddings inherently encapsulate combined semantics from all attributes and objects (e.g., ``red'', ``diamond'', ``golden'', and ``apple'').
To address this issue, End-Token-Substitution (ETS)~\cite{hu2024token} replaces attribute-rich EOS embeddings with attribute-free embeddings obtained from prompts without descriptors (e.g., ``a diamond and an apple'').
Nevertheless, ETS remains insufficient even when combined with object-wise embeddings, as the simplified EOS embeddings still aggregate semantic information across multiple objects, thereby continuing to propagate attribute leakage (see Figure~\ref{fig:limitations_existing}).

Embedding entanglement further exacerbates spatial inaccuracies in dual-branch text-based image editing frameworks.
Since dual-branch methods inject structural information from the source image into the target editing branch, entangled embeddings can produce \emph{misaligned cross-attention maps}, causing embeddings to incorrectly attend to visually similar but semantically incorrect regions during editing (Figure~\ref{fig:sub:ets}).
This visual-semantic confusion significantly exacerbates spatial inaccuracies, intensifying attribute leakage.
Although existing methods attempt to mitigate this issue through refined cross-attention alignment or explicit masking~\cite{cao2023masactrl, xu2023inversion}, these strategies fail to address the fundamental problem: EOS embeddings inherently lack spatial specificity, as they integrate semantic content from the entire prompt.
Therefore, restricting the spatial attention of EOS embeddings to specific regions is inherently ineffective, further compounding attribute leakage even when employing combined strategies (Figure~\ref{fig:sub:ets_masking}).

\begin{figure}[t!]
    \centering
    \begin{subfigure}[b]{0.24\linewidth}
        \includegraphics[width=\linewidth]{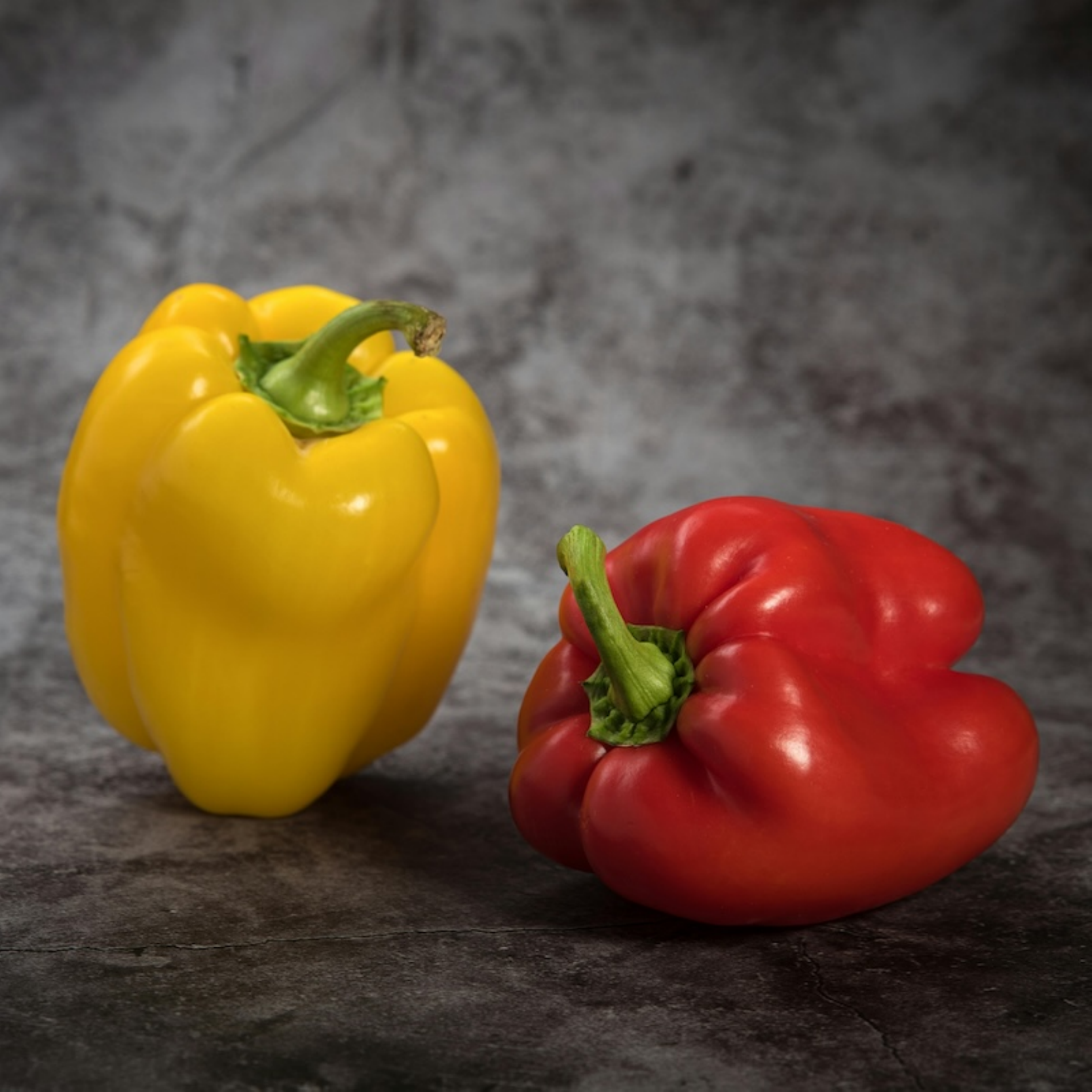}
        \caption{Source image}
    \end{subfigure}
    \begin{subfigure}[b]{0.24\linewidth}
        \includegraphics[width=\linewidth]{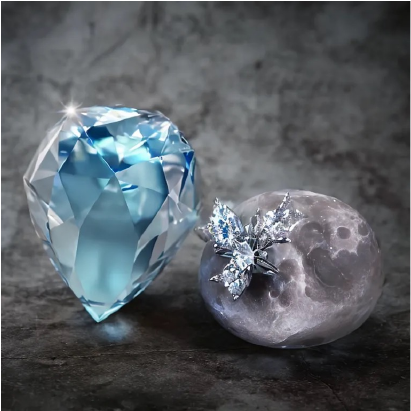}
        \caption{Original EOS}
    \end{subfigure}
    \begin{subfigure}[b]{0.24\linewidth}
        \includegraphics[width=\linewidth]{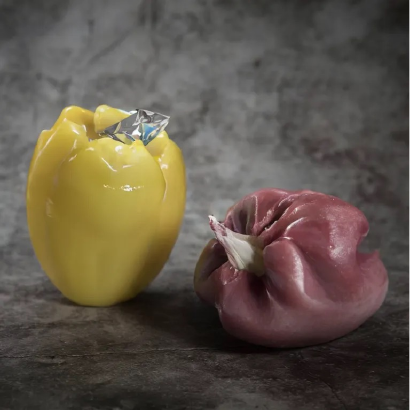}
        \caption{EOS $\xleftarrow[]{}$ $\textbf{0}$}
    \end{subfigure}
    \begin{subfigure}[b]{0.24\linewidth}
        \includegraphics[width=\linewidth]{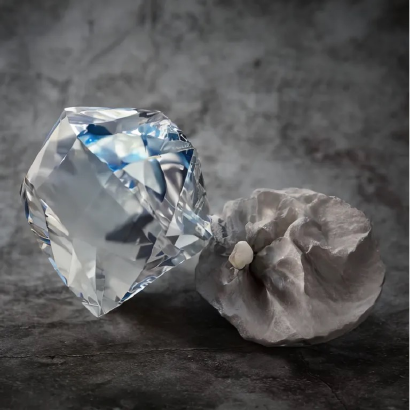}
        \caption{EOS $\xleftarrow{}$ \textbf{``''}}
    \end{subfigure}
    \caption{
    Comparison of different EOS embedding modification strategies in a single editing scenario, where ``{yellow bell pepper}'' is edited to ``{diamond}'' and ``{red bell pepper}'' to ``{moon}''. 
    (b) Uses the original EOS embeddings. 
    (c) Replaces EOS embeddings with zero vectors. 
    (d) Replaces EOS embeddings with those obtained from an empty prompt \textbf{``''}.
    }
    \label{fig:EOS_modification}
    \vspace{-0.5cm}
\end{figure}

One naive alternative could involve removing semantic content entirely from EOS embeddings by substituting them with zero vectors or embeddings derived from empty prompts.
However, as demonstrated empirically in Figure~\ref{fig:EOS_modification} and detailed in Appendix~\ref{appendix:ablation}, this simplistic solution severely degrades visual quality and editing accuracy.
These observations imply that diffusion models intrinsically depend on EOS embeddings containing semantics to achieve high-quality image editing outcomes.
Therefore, resolving attribute leakage effectively requires a dedicated strategy that carefully mitigates the unintended influence of entangled EOS embeddings without completely eliminating or overly simplifying their semantic content.
We propose such a targeted approach in the next section.

\section{Attribute-Leakage-Free Editing (ALE)}
\label{sec:method}
\begin{figure*}[ht]
    \centering
    \includegraphics[width=0.88\textwidth]{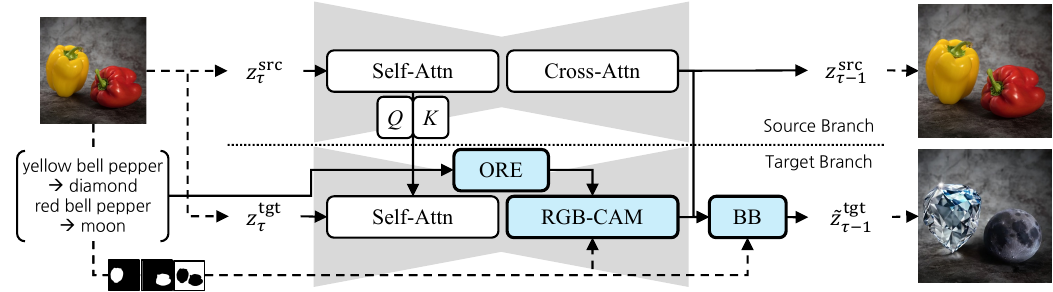}
    \caption{
        Overview of ALE. 
        The framework consists of two branches: the upper branch (source branch) processes the source latent $z_{\tau}^{\text{src}}$, and the lower branch (target branch) processes the target latent $z_{\tau}^{\text{tgt}}$ at each timestep $\tau$.
        ORE produces multiple semantically-isolated embeddings for each target object to reduce interference between unrelated objects.
        RGB-CAM refines cross-attention activations using segmentation masks, aligning the attention of each embedding to spatial regions corresponding to its target object.
        BB merges the source latent for background regions and the target latent for edited regions.
        Dashed lines indicate omitted components (e.g., the encoder, the decoder, and the segmentation model) for simplicity.
    }
    \vspace{-0.5cm}
    \label{fig:methods}
\end{figure*}

Given a source image $x^{\text{src}}$ and a list of $K$ object–level prompt pairs 
$\bigl[(y_i^{\text{src}},\,y_i^{\text{tgt}})\bigr]_{i=1}^K$, 
our goal is to generate an edited image $x^{\text{tgt}}$ that (1) replaces every $\smash{y_i^{\text{src}}}$ with $\smash{y_i^{\text{tgt}}}$ inside its designated region, (2) preserves all non-target content, and (3) avoids both TIL and TEL.

Our proposed method, ALE, is built on the dual-branch editing framework with the Denoising Diffusion Consistent Model (DDCM) virtual inversion scheme: a \emph{source branch} reconstructs $x^{\text{src}}$ from latent $\{z_\tau^{\text{src}}\}_{\tau=T}^0$ under the prompt $y^{\text{src}}_{\text{base}}\!=\!\text{``}\text{ and }\text{''}.\text{join}\bigl(\bigl[y_i^{\text{src}}\bigr]\bigr)$, 
while a \emph{target branch} denoises latent 
$\{z_\tau^{\text{tgt}}\}_{\tau=T}^0$ toward the edited image guided by 
$y^{\text{tgt}}_{\text{base}}\!=\!\text{``}\text{ and }\text{''}.\text{join}\bigl(\bigl[y_i^{\text{tgt}}\bigr]\bigr)$.
On top of this backbone we introduce three key components: ORE, 
RGB-CAM, and BB.  
Figure~\ref{fig:methods} visualizes an overall pipeline, and 
Algorithm~\ref{alg:ale_edit} lists the complete procedure.

\subsection{DDCM and Dual-Branch Editing Framework}
\label{sec:method:ddcm}
\paragraph{Virtual inversion via DDCM}\;
We adopt the Denoising Diffusion Consistent Model (DDCM) \cite{xu2023inversion}, which chooses the special variance schedule so that any noisy latent $z_\tau$ retains a closed-form link to its clean latent $z_0$ at every timestep $\tau \in [T,\dots, 0]$. 
This property enables a \emph{virtual inversion} that avoids the costly DDIM / null-text inversion. 
Because this form matches the multi-step consistency sampler of Latent Consistency Models~\cite{luo2023latent}, we can edit images in as few as 4–20 steps without explicit inversion steps.

\paragraph{Dual-branch framework}\;
Starting from an initial noise, the \emph{source branch} follows DDCM updates with ${y}_{\mathrm{base}}^{\mathrm{src}}$,
while the \emph{target branch} starts with $z_\tau^{\mathrm{tgt}}=z_\tau^{\mathrm{src}}$ and calculates the next latent $z_{\tau - 1}^{\mathrm{tgt}}$ at each step using the U-Net noise $\varepsilon_\theta(z_\tau^{\mathrm{tgt}},\tau,{y}^{\mathrm{tgt}}_{\text{base}})$ and the consistency correction term.
To preserve structure of $x^{\mathrm{src}}$, we copy the query–key tensors $({Q},{K})_{\ell,\tau}^{\mathrm{src}}$ from every self-attention layer~$\ell$ of the source branch into the corresponding layer of the target branch according to self-attention injection schedule $\mathcal{S}=\{\tau \mid T_S \le \tau \le T\}$,
where $T_S\!\in[0,T]$ controls the length of the self-attention injection schedule.
Thus, a shorter schedule ($T_s\!\approx\!T$) touches only early denoising steps and enables stronger edits, whereas a longer schedule ($T_s\!\approx\!0$) enforces stricter structural preservation.

\subsection{Object-Restricted Embeddings (ORE)}
\label{sec:method:ore}
To address leakages in prompt embedding level, ORE encodes each object prompt $\smash{y_i^{\text{tgt}}}$ in isolation, yielding a set of token-embedding matrices 
\[
E_i'=[e_{\text{BOS}},\; \underbrace{e_{\text{token}_1},\; \dots}_{\text{from tokens in }y_i^{\text{tgt}}},\; \underbrace{ e_{\text{EOS}},\; \dots}_{\text{from padded EOS tokens}}]\in \mathbb{R}^{L\times d}
\]
where \(L\) is the padded prompt length and \(d\) the embedding dimension. 
For example, in Figure~\ref{fig:limitations_existing} case, $E'_1$ is obtained as $[e_{\text{BOS}}, \; e_{\text{a}}, \; e_{\text{red}}, \; e_{\text{diamond}}, \; e_{\text{EOS}}, \; \dots]$.
And we construct a base embedding $E'_{\text{base}}$, by encoding $y_{\mathrm{base}}^{\mathrm{tgt}}$ and splicing $\smash{E'_i}[y_i^{\text{tgt}}]$ back into their original spans, to calculate base value tensor in RGB-CAM. 
Because no token embedding in $\{E'_i\}$ can influence another object’s span and EOS embeddings in $E'_i$ only contain semantics of $y_i^{\text{tgt}}$, subsequent cross-attention receives \emph{semantically disentangled} embeddings, thereby preventing leakage at its source. 

\subsection{Region-Guided Blending for Cross-Attention Masking (RGB-CAM)}
\label{sec:method:rgbcam}
Standard cross-attention layers in diffusion U-Net accept a \emph{single} value tensor $V$ and thus cannot exploit multiple OREs.  
RGB-CAM replaces the vanilla cross-attention output with a spatially
blended tensor
\[
    A \;=\;\sum_{i=1}^{K} (M\odot m_i)\,\,V_i \;+\; (M\odot m_{\text{back}})\,\,V_{\text{base}},
\]
where  
$M=\text{attention\_map}(Q,K)$ is the base cross attention map,  
$V_i=W_v(E'_i)$, $K=W_k(\text{Encoder}_{\text{text}}(y_{\text{base}}^{\text{tgt}}))$, $V_{\text{base}}=W_v(E'_{\text{base}})$, and  
$\{m_i\}$, $m_{\text{back}}$ are object and background segmentation masks from Grounded-SAM~\cite{ren2024grounded}.
Since masks are not pixel-perfect, we apply a slight dilation.
The masked tensors $(M\odot m_i)V_i$ \emph{localize} each ORE to its designated region, eliminating target-internal leakage, while the background term preserves areas outside all masks.
Note that only when ORE and RGB-CAM operate \emph{in tandem} does ALE produce leakage-free results.

\subsection{Background Blending (BB)}
\label{sec:method:bb}
Even with perfect cross-attention, backgrounds remain weakly constrained because $\{y^{\text{tgt}}_i\}$ mention only target objects.  
At every timestep $\tau$ we blend the source latent by the background mask as a final step:
\[
    \tilde z_{\tau}^{\text{tgt}}
    \;=\;
    m_{\text{back}}\odot z_{\tau}^{\text{src}}
    \;+\;
    (1-m_{\text{back}})\odot z_{\tau}^{\text{tgt}}.
\]
BB guarantees preservation of non-edited regions, suppressing TEL without expensive threshold tuning required by
prior local-blending heuristics~\cite{hertz2022prompt}.

\begin{algorithm}[t]
\caption{Attribute-Leakage-Free Editing}
\label{alg:ale_edit}

\KwIn{Source image $x^{\text{src}}$, object-level prompt pairs $\bigl[(y_i^{\text{src}},\,y_i^{\text{tgt}})\bigr]_{i=1}^K$, self-attention injection schedule $\mathcal{S}=\{\tau\mid 0\le\tau\le T_S\}$}
\KwOut{Edited image $x^{\text{tgt}}$}

\medskip
\textbf{Pre-processing}

\Indp
Form concatenated base prompts $y^{\text{src}}_{\text{base}},\;y^{\text{tgt}}_{\text{base}}$\;
  
Encode object-restricted embeddings $\{E'_i\}_{i=1}^K$ and the base embedding $E'_{\text{base}}$ (ORE)\;
  
Obtain object masks $\{m_i\}_{i=1}^K$ and background mask $m_{\text{back}}$ with Grounded-SAM\;
\Indm

\textbf{Initialization}

\Indp
Sample initial noise $z_T^{\text{src}}\!\sim\!\mathcal{N}(0,I)$\;
  
Set $z_T^{\text{tgt}}\!\leftarrow\!z_T^{\text{src}}$\;
\Indm



\For{$\tau=T$ \KwTo $1$}{
    {\color{gray}\tcp*[h]{Source branch}}
    
    Predict noise $\hat\varepsilon_\tau^{\text{src}}\!\leftarrow\!\varepsilon_\theta\bigl(z_\tau^{\text{src}},\tau,y_{\text{base}}^{\text{src}}\bigr)$\;
    
    Update $z_{\tau-1}^{\text{src}}$ with DDCM sampling\;

    {\color{gray}\tcp*[h]{Target branch}}
    
    \lIf{$\tau\in\mathcal{S}$}{copy self-attention $Q,K$ tensors from the source branch}
    
    Predict noise $\hat\varepsilon_\tau^{\text{tgt}}\!\leftarrow\!\varepsilon_\theta\bigl(
        z_\tau^{\text{tgt}},\tau,y_{\text{base}}^{\text{tgt}};\;
        \text{RGB-CAM}\!\left[\{E'_i,m_i\},E'_{\text{base}},m_{\text{back}}\right]
    \bigr)$\;
    Update $z_{\tau-1}^{\text{tgt}}$ with DDCM sampling\;
    
    {\color{gray}\tcp*[h]{Background blending (BB)}}
    $z_{\tau-1}^{\text{tgt}}\!\leftarrow\!
    m_{\text{back}}\!\odot\! z_{\tau-1}^{\text{src}}
    \;+\;
    \bigl(1-m_{\text{back}}\bigr)\!\odot\! z_{\tau-1}^{\text{tgt}}$\;
}
$x^{\text{tgt}}\!\leftarrow\!\mathrm{Decoder}\!\bigl(z_{0}^{\text{tgt}}\bigr)$\;

\KwRet{$x^{\text{tgt}}$}

\end{algorithm}

\section{Experiments}
\label{sec:exp}
%


\begin{figure*}[ht]
    \centering
    \includegraphics[width=\textwidth]{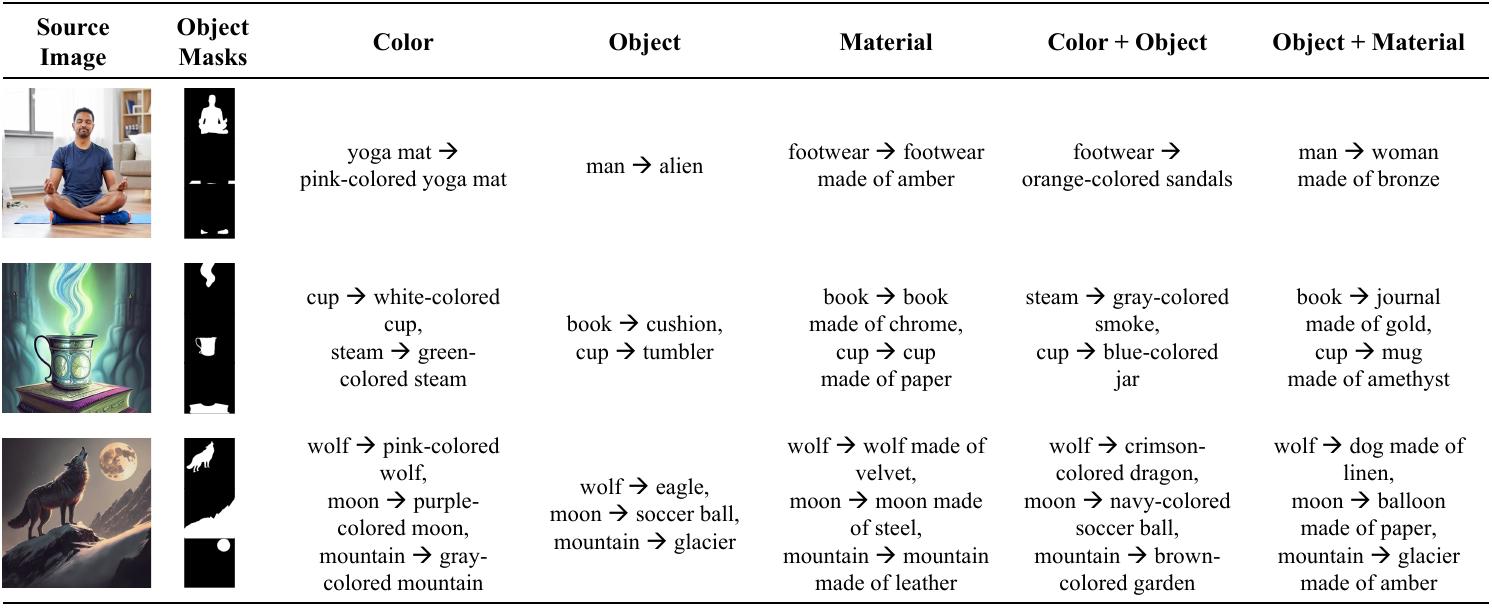}
    \vspace{-0.7cm}
     \caption{
     Examples of \benchmark{}.
     Source images are shown with object binary masks.
     Edit types are color, object, material modifications, and their combinations.
     Rows depict 1-, 2-, and 3-object edits, respectively.
    }
    \label{fig:benchmark_examples}
    \vspace{-0.3cm}
\end{figure*}

\subsection{Experiment Setup}
\label{sec:exp:setup}

\paragraph{\benchmark{} construction}
\label{sec:exp:benchmark}
Prior benchmarks for text-guided image editing~\cite{ju2023direct,chakrabarty2024lomoe} focus on the visual quality of the edited result itself and neglect \emph{attribute leakage}.
Although they measure background preservation metrics that are similar to TELS, they overlook TIL.
This makes it difficult to analyze how well a method achieves precise editing.  
To fill this gap, we introduce Attribute-Leakage-Evaluation Benchmark (\benchmark{}), a dedicated testbed for attribute leakage evaluation.  
\benchmark{} systematically varies the number of objects to be edited and the editing type, grouped into five categories:  
(1) \textit{color}, (2) \textit{object}, (3) \textit{material}, (4) \textit{color + object}, and (5) \textit{object + material}.  
Each source image is paired with multiple prompts for each edit type, enabling fine-grained analysis across diverse scenarios. 
Figure~\ref{fig:benchmark_examples} shows examples of \benchmark{}.
Full construction details and dataset statistics are provided in Appendix~\ref{appendix:benchmark}.

\paragraph{Evaluation Metrics}
\label{sec:exp:metric}
We evaluate image-editing performance using the following metrics:
\begin{itemize}
    \item \emph{Structure Distance}~\cite{tumanyan2022splicing} quantifies how well the edited image ${x}^{\text{tgt}}$ preserves the spatial layout  of ${x}^{\text{src}}$.  
    A lower score indicates better structural consistency.
    \item  \emph{Editing Performance} is measured by the cosine similarity between the CLIP embeddings of $x^{\text{tgt}}$ and prompt $y^{\text{tgt}}$.  
    Higher similarity reflects more faithful edits.
    \item \emph{Background Preservation} is assessed on the non-edited regions using PSNR, SSIM~\cite{wang2004image}, LPIPS~\cite{zhang2018unreasonable}, and MSE between ${x}^{\text{tgt}}$ and ${x}^{\text{src}}$.  
  Higher PSNR/SSIM and lower LPIPS/MSE signify better preservation.
\end{itemize}

Furthermore, we introduce new metrics: 
\begin{itemize}   
    \item \emph{Target-Internal Leakage Score (TILS)} measures unintended modifications inside other target-object regions as follows:
    \[
        \text{TILS} = \frac{1}{K(K-1)}\sum_{i \not= j}^{K}
        \text{CLIP} \left(
        x_{\text{tgt}} \odot m_j, y_i^{\text{tgt}}
        \right)
        \label{eq/ti},
    \]
    where CLIP represents the CLIP similarity score, $K$ is the number of objects to be edited, $x^{\text{tgt}}$ is the edited image, ${m}_j$ is the $j$-th object mask, and $y_i^{\text{tgt}}$ is the target prompt for $i$-th object.
    A lower TILS implies that, as the user intended, the target objects did not affect each other. 

    \item \emph{Target-External Leakage Score (TELS)} measures unintended changes in the background (non-edited regions) as follows:
    \[
        \text{TELS} = \frac{1}{K}\sum_{i=1}^{K}
        \text{CLIP}\left(
            x^{\text{tgt}} \odot \left( \textbf{1} - \bigcup_{j=1}^{K}{m}_j\right), y_i^{\text{tgt}} 
        \right).
        \label{eq/te}
    \]
    The mean CLIP scores between the background and each target prompt are computed for multiple object edits.
    A lower TELS indicates minimal TEL, which ensures that the outside of the targets remains unchanged.
\end{itemize}

\paragraph{Baselines}
\label{sec:exp:baseline}
For comparison, we selected tuning-free image editing methods including Prompt-to-Prompt (P2P)~\cite{hertz2022prompt}, MasaCtrl~\cite{cao2023masactrl}, Free-Prompt-Editing (FPE)~\cite{zou2024towards}, and InfEdit~\cite{xu2023inversion} (see Appendix~\ref{sec:related work}, \ref{appendix:exp_details} for further details).

\begin{figure*}[ht]
    \centering
    \resizebox{\textwidth}{!}{%
        \includegraphics[]{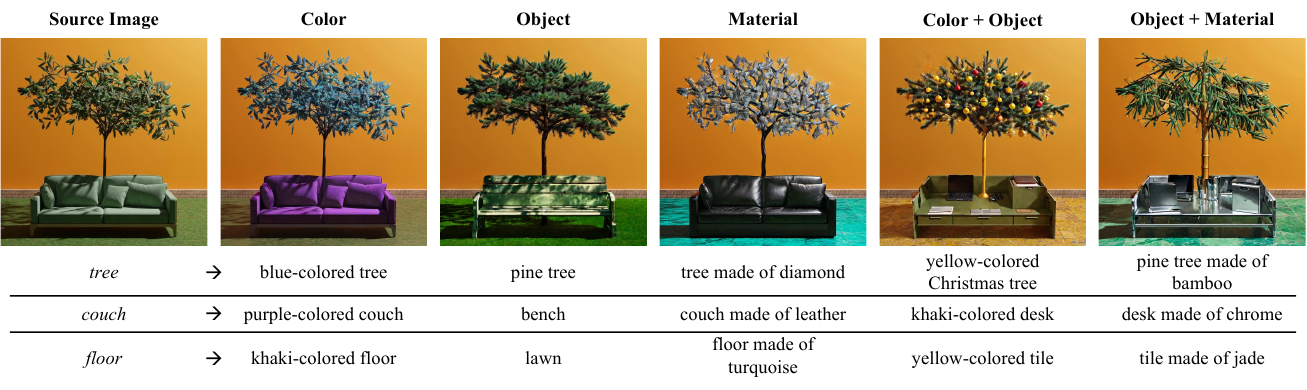}
    }
    \vspace{-0.75cm}
     \caption{
     Qualitative examples of ALE applied to three source objects (\textit{tree}, \textit{couch}, and \textit{floor}) under different editing types.
     The bottom rows show the corresponding prompts describing each transformation.
    }
    \label{fig:qualitative}
    \vspace{-0.3cm}
\end{figure*}

\begin{table*}[th]\centering

\resizebox{\textwidth}{!}{%
\begin{tabular}{cccccccccccccc}
\toprule
\multirow{2}{*}{\textbf{Method}} &
&
\multirow{2}{*}{\textbf{TELS ↓}}
 &
 &
 \multirow{2}{*}{\textbf{TILS ↓}} &
 &
\multirow{2}{*}{\textbf{\begin{tabular}[c]{@{}c@{}}Structure \\ Distance ↓\end{tabular}}}
&
&
\multirow{2}{*}{\textbf{\begin{tabular}[c]{@{}c@{}}Editing \\ Performance ↑\end{tabular}}}
 && \multicolumn{4}{c}{\textbf{\begin{tabular}[c]{@{}c@{}}Background  Preservation\end{tabular}}} \\
 && & & & & & & & & PSNR ↑& LPIPS ↓ & MSE ↓& SSIM ↑\\ \cline{1-1} \cline{3-3} \cline{5-5} \cline{7-7} \cline{9-9} \cline{11-14} 

P2P      &  & 21.52          &           & 17.26          &           & 0.1514          &           & 20.67          &           & 11.15          & 0.4495          & 0.0879          & 0.5589          \\
MasaCtrl &  & 20.18          &           & 16.74          &           & 0.0929          &           & 20.01          &           & 14.99          & 0.2930          & 0.0418          & 0.7346          \\
FPE      &  & 21.07          &           & 17.38          &           & 0.1164          &           & 21.89          &           & 12.82          & 0.3903          & 0.0656          & 0.6052          \\
InfEdit  &  & 19.59          &           & 16.69          &           & 0.0484          &           & 21.78          &           & 16.74          & 0.2034          & 0.0340          & 0.7709          \\
\cline{1-1} \cline{3-3} \cline{5-5} \cline{7-7} \cline{9-9} \cline{11-14} 
ALE     &  & \textbf{16.03} & & \textbf{15.28} & & \textbf{0.0167} &  & \textbf{22.20} & & \textbf{30.04} & \textbf{0.0361} & \textbf{0.0014} & \textbf{0.9228} \\ \bottomrule

\end{tabular}
}
\vspace{-0.1cm}
\caption{
Comparison of editing performances on \benchmark{}.
ALE demonstrates the lowest attribute leakage, highest structure preservation, and superior editing performance, indicating a more precise and controlled editing.
}
\label{table:main_results}

\end{table*}
\begin{table*}[th]\centering

\resizebox{\textwidth}{!}{%
\begin{tabular}{cccccccccccccc}
\toprule
\multirow{2}{*}{\textbf{\begin{tabular}[c]{@{}c@{}}\# of Editing\\ Objects\end{tabular}}} && 
\multirow{2}{*}{\textbf{TELS ↓}}
 &
 &
 \multirow{2}{*}{\textbf{TILS ↓}} &
 &
\multirow{2}{*}{\textbf{\begin{tabular}[c]{@{}c@{}}Structure \\ Distance ↓\end{tabular}}}
&
&
\multirow{2}{*}{\textbf{\begin{tabular}[c]{@{}c@{}}Editing \\ Performance ↑\end{tabular}}}
 && \multicolumn{4}{c}{\textbf{\begin{tabular}[c]{@{}c@{}}Background  Preservation\end{tabular}}} \\
 && & & & & & & & & PSNR ↑& LPIPS ↓ & MSE ↓& SSIM ↑\\ \cline{1-1} \cline{3-3} \cline{5-5} \cline{7-7} \cline{9-9} \cline{11-14} 
 
1 &  & 16.41 &  & -   &  & 0.00876 &  & 22.62 &  & 30.01 & 0.0405 & 0.00167 & 0.9049 \\
2 &  & 16.00 &  & 15.42 &  & 0.01648 &  & 22.06 &  & 30.06 & 0.0360 & 0.00146 & 0.9235 \\
3 &  & 15.89 &  & 15.36 &  & 0.02460 &  & 22.19 &  & 30.01 & 0.0323 & 0.00154 & 0.9426 \\
\bottomrule
\end{tabular}
}
\vspace{-0.1cm}
\caption{
Performance of ALE on \benchmark{} based on the number of objects edited.
Our method maintains low attribute leakage and strong background preservation even as the number of editing objects increases.
}
\label{table:per_objects_results}
\end{table*}
\begin{table*}[ht!]\centering

\resizebox{\textwidth}{!}{%
\begin{tabular}{cccccccccccccc}
\toprule
\multirow{2}{*}{\textbf{\begin{tabular}[c]{@{}c@{}}Editing\\ Type\end{tabular}}} && 
\multirow{2}{*}{\textbf{TELS ↓}}
 &
 &
 \multirow{2}{*}{\textbf{TILS ↓}} &
 &
\multirow{2}{*}{\textbf{\begin{tabular}[c]{@{}c@{}}Structure \\ Distance ↓\end{tabular}}}
&
&
\multirow{2}{*}{\textbf{\begin{tabular}[c]{@{}c@{}}Editing \\ Performance ↑\end{tabular}}}
 && \multicolumn{4}{c}{\textbf{\begin{tabular}[c]{@{}c@{}}Background  Preservation\end{tabular}}} \\
 && & & & & & & & & PSNR ↑& LPIPS ↓ & MSE ↓& SSIM ↑\\ \cline{1-1} \cline{3-3} \cline{5-5} \cline{7-7} \cline{9-9} \cline{11-14} 

 
Color           &  & 17.63 &  & 16.21 &  & 0.00890 &  & 23.12 &  & 32.97 & 0.0288 & 0.00079 & 0.9309 \\
Material        &  & 17.15 &  & 15.96 &  & 0.01179 &  & 22.94 &  & 30.63 & 0.0339 & 0.00120 & 0.9248 \\
Object          &  & 15.86 &  & 16.25 &  & 0.01974 &  & 21.82 &  & 29.03 & 0.0386 & 0.00182 & 0.9218 \\
Color+Object    &  & 15.30 &  & 14.01 &  & 0.02306 &  & 22.15 &  & 28.60 & 0.0407 & 0.00205 & 0.9206 \\
Object+Material &  & 14.55 &  & 14.51 &  & 0.01956 &  & 21.42 &  & 28.88 & 0.0393 & 0.00191 & 0.9201 \\
\bottomrule

\end{tabular}
}
\vspace{-0.1cm}
\caption{
Performance of ALE on \benchmark{} across various editing types (color, object, material, and combinations).
The results show consistent low attribute leakage and high editing performance.
}
\label{table:results_per_type}
\end{table*}

\subsection{Main Results}
\label{sec:exp:main}
ALE outperforms existing methods in both mitigating attribute leakage and producing high-quality edits, as seen in Table~\ref{table:main_results} and Figure~\ref{fig:qualitative}.
In particular, ALE achieves the lowest TELS and TILS, reflecting its ability to precisely apply attributes solely to the designated target regions. 
Across different numbers of editing objects (Table~\ref{table:per_objects_results}) and different editing types (Table~\ref{table:results_per_type}), ALE demonstrates its robust performance.
Figure~\ref{fig:qualitative} illustrates qualitative examples of ALE for various editing types.
These results indicate that our method can effectively address both TEL and TIL.
Detailed results are in Appendix~\ref{appendix:results}.
Furthermore, the quantitative and qualitative results on PIE-Bench are in Appendix~\ref{appendix:ablation}.

\begin{figure}[!ht]
    \centering
    \begin{subfigure}[b]{0.24\linewidth}
        \includegraphics[width=\linewidth]{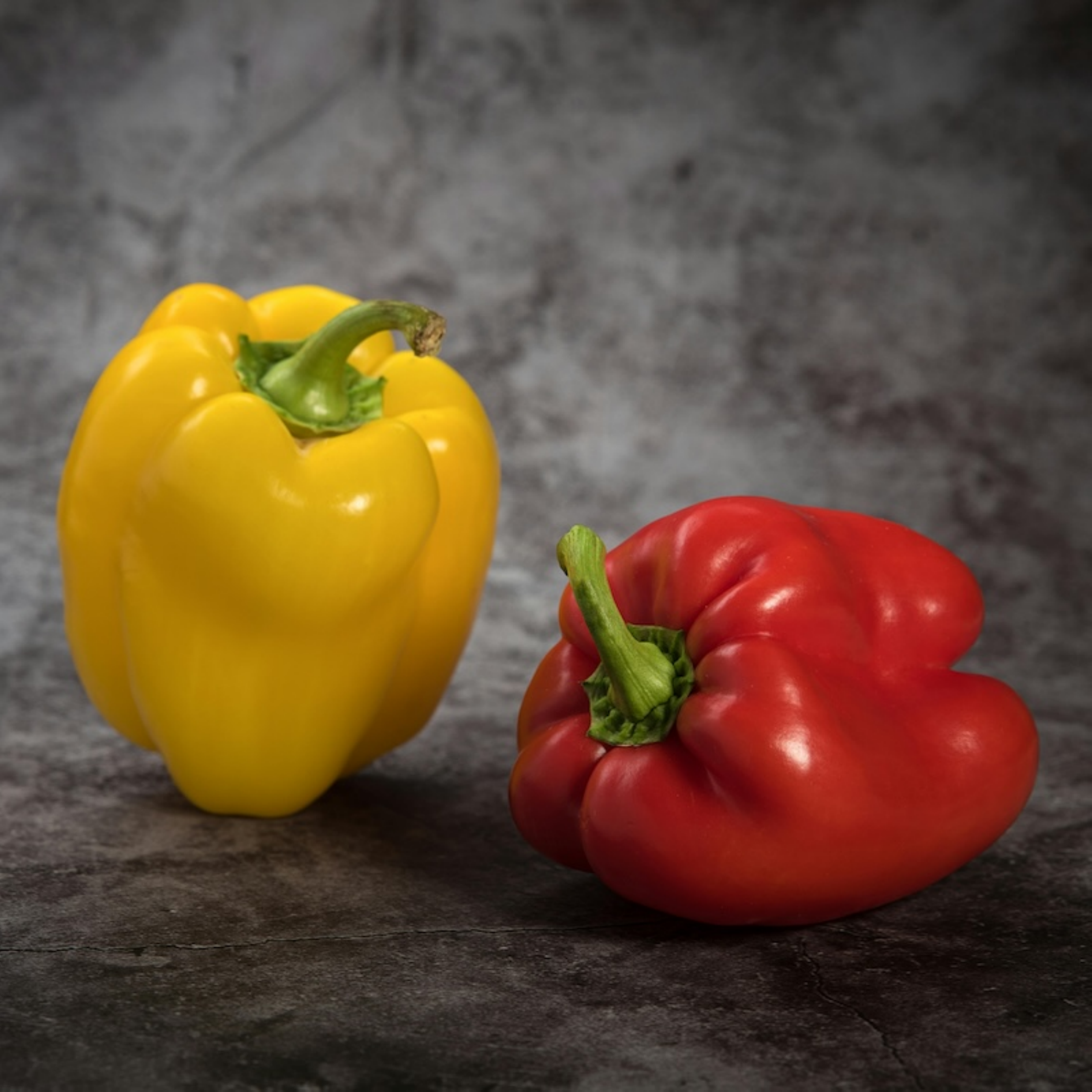}
        \caption{Source image}
    \end{subfigure}
    \begin{subfigure}[b]{0.24\linewidth}
        \includegraphics[width=\linewidth]{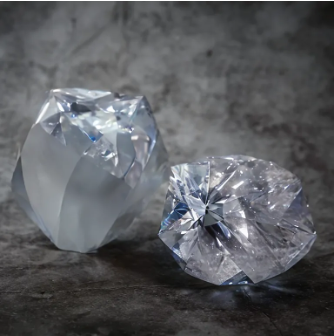}
        \caption{BB}
        \label{fig:ablation:bb}
    \end{subfigure}
    \begin{subfigure}[b]{0.24\linewidth}
        \includegraphics[width=\linewidth]{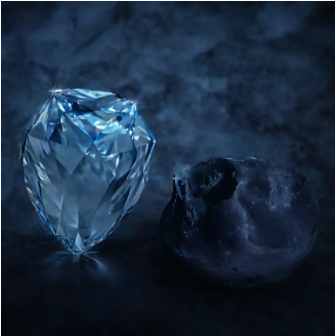}
        \caption{ORE + RGB}
        \label{fig:ablation:ore_rgb}
    \end{subfigure}
    \begin{subfigure}[b]{0.24\linewidth}
        \includegraphics[width=\linewidth]{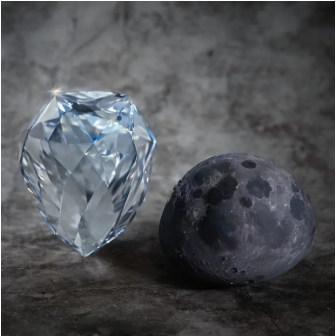}
        \caption{ALE}
        \label{fig:ablation:ale}
    \end{subfigure}
    \vspace{-0.1cm}
     \caption{
        Qualitative ablation results for the editing: {yellow bell pepper} $\rightarrow$ {diamond}, {red bell pepper} $\rightarrow$ {moon}. 
        (b) BB resolves TEL but not TIL. 
        (c) ORE + RGB-CAM reduces TIL but not TEL. 
        (d) ALE (BB + ORE + RGB-CAM) shows no TIL and TEL.
    }
    \label{fig:ablation}
\end{figure}

\subsection{Ablation Studies}
\label{sec:exp:ablation}
We present ablation results on BB, ORE, and RGB-CAM in Figure~\ref{fig:ablation}.
Each module plays a complementary role in preventing attribute leakage.
Using only BB (Figure~\ref{fig:ablation:bb}) helps preserve background regions and suppresses TEL, but fails to prevent TIL.
On the other hand, ORE + RGB-CAM (Figure~\ref{fig:ablation:ore_rgb}) reduces TIL by disentangling text embeddings and aligned cross-attention. 
However, without BB, it cannot preserve the original background, and TEL remains.
Only when all three modules—ORE, RGB-CAM, and BB—are used together (Figure~\ref{fig:ablation:ale}), ALE achieves precise and leakage-free editing.
Qualitative results and more examples are provided in Appendix~\ref{appendix:ablation}.

\section{Limitations}
\label{sec:limiations}
While ALE-Bench provides a focused framework for evaluating attribute leakage, and ALE achieves strong performance on \emph{rigid} attribute edits, both are currently limited to local and relatively simple transformations—such as changes in color, object identity, or material. 
They do not support or evaluate non-rigid transformations like style transfer, pose changes, or adding/deleting objects, where defining and detecting attribute leakage becomes ambiguous. 
This focus reflects our aim to establish a clear and measurable foundation before addressing more complex editing scenarios. 
Although the benchmark offers $3,000$ detailed editing scenarios across 20 carefully curated images, the small image set may limit how well results generalize to models trained on larger or more diverse datasets.
Future work could expand both the editing model and benchmark to support richer transformations and broader datasets.
More limitations are discussed in Appendix~\ref{appendix:limitations}.

\section{Conclusion}
\label{sec:conclusion}

In this paper, we addressed the issue of attribute leakage in diffusion-based image editing, 
focusing on two types of leakage: Target-External Leakage (TEL)—unintended edits in non-target regions—and 
Target-Internal Leakage (TIL)—interference among attributes of different targets.
To mitigate these problems, we introduced Attribute-Leakage-Free Editing (ALE), a tuning-free framework that combines three key components:
Object-Restricted Embeddings (ORE), which localize attribute semantics to each target object embedding;
Region-Guided Blending for Cross-Attention Masking (RGB-CAM), which constrains cross-attention with segmentation masks to avoid unintended inter-object attribute mixing; 
and Background Blending (BB), which preserves the source image in backgrounds.

We also presented \benchmark{}, a dedicated benchmark for rigorously evaluating attribute leakage across diverse multi-object editing scenarios. 
\benchmark{} introduces new quantitative metrics—TELS and TILS—that effectively quantify unintended modifications, providing comprehensive measures to assess editing fidelity and consistency.

Our extensive experimental validation demonstrated that ALE significantly outperforms existing tuning-free editing methods, achieving state-of-the-art performance by effectively minimizing attribute leakage while maintaining high editing quality and structural consistency. 
By effectively addressing attribute leakage with minimal computational overhead, ALE enhances the reliability and precision of multi-object image editing tasks.



\clearpage

\section*{Acknowledgements}




This work was partly supported by Institute of Information \& Communications Technology Planning \& Evaluation (IITP) and IITP-ITRC (Information Technology Research Center) grant funded by the Korea government (MSIT) (No.RS-2019-II191906, Artificial Intelligence Graduate School Program (POSTECH); No.RS-2021-II212068, Artificial Intelligence Innovation Hub; IITP-2025-00437866; RS-2024-00509258, Global AI Frontier Lab) and partly supported by Seoul R\&BD Program through the Seoul Business Agency (SBA) funded by The Seoul Metropolitan Government (SP240008).

{
    \small
    \bibliographystyle{ieeenat_fullname}
    \bibliography{main}
}

\appendix
\clearpage

\section{Related Work}
\label{sec:related work}

\subsection{Diffusion-based Image Editing}
Diffusion-based image editing strives to (1) preserve the visual content of a given source image while (2) modifying specific regions as instructed by text prompts.
Prompt-to-Prompt (P2P)~\cite{hertz2022prompt} pioneered this line of research by introducing \emph{cross-attention scheduling}, which injects the cross-attention maps obtained during the reconstruction of the source image back into the editing process.
Subsequent works further refined attention manipulation:
MasaCtrl~\cite{cao2023masactrl} imposes \emph{mutual self-attention control} to maintain spatial consistency, whereas Free-Prompt-Editing (FPE)~\cite{liu2024towards} decomposes cross-attention scheduling layer by layer for finer control.

A complementary thread focuses on \emph{inversion}, operating under the intuition that an accurate inversion of the source image yields higher-quality edits.
Several methods optimize text embeddings during inversion~\cite{mokady2023null,yang2023dynamic,li2023stylediffusion,dong2023prompt,xu2023inversion}; among them, InfEdit~\cite{xu2023inversion} proposes a training-free \emph{Virtual Inversion} technique that achieves state-of-the-art results on multiple benchmarks.

Despite these advances, \emph{multi-object} image editing remains under-explored.
ZRIS~\cite{yu2023zero} handles multi-object cases, but segments objects for \emph{referring image segmentation}, not editing.
Editing multiple target objects sequentially is straightforward but computationally expensive, as each object requires a separate diffusion pass.
Our work addresses this gap by proposing a training-free framework that simultaneously handles multiple editing prompts and aligns user intent with the attention mechanism, enabling efficient multi-object edits without sacrificing quality.


\subsection{Mitigating Attribute Leakage}
\emph{Attribute leakage} occurs when a diffusion model assigns an attribute to an unintended object.
Early work addressed the problem by injecting explicit linguistic structure:  
StructureDiffusion Guidance~\cite{feng2022training} constrains generation with a constituency tree or scene graph, while Attend‐and‐Excite~\cite{chefer2023attend} and SynGen~\cite{rassin2023linguistic} refine cross‐attention so that each word attends to a single spatial region.  
These methods focus on the \emph{attention maps} themselves, yet leakage can also stem from \emph{text embeddings}: even a perfect attention map fails if the prompt embedding is semantically entangled.

ToMe~\cite{hu2024token} tackles embedding‐level entanglement via \emph{End Token Substitution} (ETS).  
It replaces the EOS embedding of the full prompt (e.g., ``a yellow cat and a white dog'') with the EOS embedding of a stripped prompt that omits attributes (e.g., ``a cat and a dog''), thereby suppressing color–attribute leakage (``yellow dog'', ``white cat'').  
However, ETS does not address noun‐to‐noun confusion (``cat'' versus ``dog'') and, being designed for pure image generation, offers no guarantee of consistency with a given source image—an essential requirement for editing.
DPL~\cite{yang2023dynamic} reduces leakage by iteratively optimizing token embeddings at inference time to align cross‐attention maps with the prompt, yet this costly optimization still leaves leakage when EOS embeddings remain entangled (see Figure~\ref{fig:limitations_existing}).  

In summary, existing approaches either leave EOS embeddings untouched or require high-cost optimization. 
Our method instead offers a lightweight, \emph{optimization‐free} pipeline that simultaneously disentangles embeddings and aligns attention, achieving lower attribute leakage while preserving faithfulness to the source image.

\section{Benchmark Construction Details}
\label{appendix:benchmark}

\paragraph{Benchmark overview}
Our benchmark is designed to evaluate attribute leakage in image editing tasks using diffusion models.
Unlike existing benchmarks that focus on image quality and background preservation, our benchmark emphasizes preventing unintended changes in both target-external and target-internal regions.
It consists of 20 diverse images, semi-automated object masks, and succinct prompt pairs for various editing types.
To comprehensively evaluate models, we generate 10 random edit prompts for each combination of 5 edit types and 1–3 edited objects per image, resulting in a total of 3,000 diverse editing scenarios.
By covering diverse editing scenarios and offering precise evaluation metrics, our benchmark provides a robust framework for improving the precision of image editing methods.
Figure \ref{fig:benchmark_examples} illustrates examples, showing the source images, object masks, and associated editing prompts.

\paragraph{Image selection}
We curated a dataset of 20 images, evenly split between natural and artificial scenes, to provide diverse and challenging editing scenarios.
All images were drawn from both free image repositories and the PIE-bench dataset \cite{xu2023inversion}.
To ensure complexity, we included only images containing at least three distinct objects.

\paragraph{Prompt construction}
\benchmark{} provides five editing types. 
The prompt templates for different editing types are as follows:
\begin{enumerate}
    \item Color change: ``\{color\}-colored \{object\}'' (e.g., ``{car}'' $\xrightarrow{}$ ``{red-colored} car'').
    \item Object change: ``\{new object\}'' (e.g., ``{car}'' $\xrightarrow{}$ ``{bus}'').
    \item Material change: ``\{object\} made of \{material\}'' (e.g., ``{car}'' $\xrightarrow{}$ ``car {made of gold}'').
    \item Color and object change:  ``\{color\}-colored \{new object\}" (e.g., ``{car}'' $\xrightarrow{}$ ``{blue-colored bus}'').
    \item Object and material change: ``\{new object\} made of \{material\}'' (e.g., ``{car}'' $\xrightarrow{}$ ``{bus made of gold}'').
\end{enumerate}
We intentionally excluded combinations like ``color and material'' and ``color, object and material'' because such cases often lead to unrealistic or ambiguous prompts, such as ``\textbf{silver}-colored car made of \textbf{gold}''.
These kinds of descriptions are inherently challenging to interpret or generate, even for a human, making them impractical editing scenarios.

For each image, we generated 10 unique and random edit prompt instances for every combination of edit type and number of objects to edit.
These prompts were created using attribute dictionaries containing target instances for colors, objects, and materials, with the assistance of ChatGPT to ensure diversity and consistency.
This approach results in a systematic exploration of the attribute space across 20 images, 5 edit types, and varying numbers of objects, covering a total of 3,000 unique editing scenarios.
Additionally, we emphasize the importance of user convenience by designing minimal prompt pairs that specify only the intended modification, avoiding the verbosity commonly seen in previous benchmarks.

\paragraph{Evaluation metrics}
In addition to standard metrics from PIE-bench—such as structural distance, background preservation (PSNR, SSIM, LPIPS, MSE), and editing performance (CLIP similarity)—we propose two novel metrics specifically designed to evaluate attribute leakage.
The Target-External-Leakage Score (TELS) metric quantifies unintended changes to background regions during editing.
This is calculated by measuring the CLIP scores between the background regions of the edited image and the target prompt.
Lower TELS indicate minimal impact on the background, ensuring that non-target regions remain unaffected.
The Target-Internal-Leakage Score (TILS) metric captures unintended cross-influence between multiple edited objects.
For each edited object, we compute the CLIP scores between its edited region and the prompts intended for other objects, then take the mean scores across all object pairs.
Lower TILS indicate that edits are confined to their respective objects without unintended interactions or overlaps.

\paragraph{Comparison with LoMOE-Bench}
LoMOE-Bench~\cite{chakrabarty2024lomoe} evaluates overall fidelity in multi-object editing using approximately 1k edits across 64 images. In contrast, \benchmark{} focuses on probing \emph{attribute leakage}, generating \emph{3k edits} from just 20 carefully selected images. Rather than scaling the dataset broadly, \benchmark{} emphasizes depth by designing diverse, leakage-prone scenarios for each image. Since each additional image requires new object masks and source–target prompt pairs, annotation costs grow linearly. As a result, the two benchmarks serve complementary purposes: LoMOE-Bench measures broad editing fidelity, while ALE-Bench targets leakage robustness.

\section{Experiments Details}
\label{appendix:exp_details}

\paragraph{Prompt construction}
For methods such as MasaCtrl and FPE that require only a single target prompt, the target prompt was constructed by concatenating all target object prompts with ``and'' to form a prompt.
For methods like P2P and InfEdit that require both a source and a target prompt, the source prompt was similarly created by concatenating the source object prompts, while the target prompt was constructed by concatenating the target object prompts.

\paragraph{Hyperparameters}
\label{appendix:hyperparameters}
For our method, we set the inference steps to 15 and the mask dilation ratio to 0.01, corresponding to a dilation of seven pixels.
The self-attention control schedule was adjusted according to the type of edit: 
1.0 for colors, 0.5 for objects, color+object, and material+object, and 0.6 for material.
The same self-attention control schedule was applied to InfEdit and P2P, as this hyperparameter is shared.
For all other hyperparameters of the baseline methods (MasaCtrl, FPE, P2P, InfEdit), we used the default settings provided in their official implementations.


\section{Additional Results}
\label{appendix:results}

\paragraph{Runtime comparison}
As shown in Table~\ref{tab:runtime}, ALE and InfEdit achieve significantly faster runtimes compared to other baselines, requiring only a few seconds per edit. This efficiency comes from leveraging virtual inversion via DDCM. In contrast, methods like P2P, MasaCtrl, and FPE rely on more expensive DDIM or null-text inversion processes, resulting in runtimes of nearly one minute per edit.

\begin{table}[t!]
    \vspace{-0.1cm}
    \centering
    \resizebox{\linewidth}{!}{
    \begin{tabular}{cccccc}
    \toprule
    Method & P2P & MasaCtrl & FPE & InfEdit & ALE (Ours) \\
    \midrule
    Runtime (sec) ↓& 61.2 & 63.6 & 50.9 & 5.41 & \textbf{4.31} \\
    \bottomrule
    \end{tabular}
    }
    \caption{Average runtime per edit on an RTX 6000 Ada Gen.}
    \label{tab:runtime}
\end{table}

\paragraph{By object count}
\begin{table*}[p]\centering

\resizebox{\textwidth}{!}{%
\begin{tabular}{cccccccccccccc}
\toprule
\multirow{2}{*}{\textbf{Method}} &
&
\multirow{2}{*}{\textbf{TELS ↓}}
 &
 &
\multirow{2}{*}{\textbf{\begin{tabular}[c]{@{}c@{}}Structure \\ Distance ↓\end{tabular}}}
&
&
\multirow{2}{*}{\textbf{\begin{tabular}[c]{@{}c@{}}Editing \\ Performance ↑\end{tabular}}}
 && \multicolumn{4}{c}{\textbf{\begin{tabular}[c]{@{}c@{}}Background  Preservation\end{tabular}}} \\
 && & & & & & & PSNR ↑& LPIPS ↓ & MSE ↓& SSIM ↑\\
 \cline{1-1} \cline{3-3} \cline{5-5} \cline{7-7} \cline{9-12} 


P2P      &  & 24.91          &           & 0.1513           &           & 21.53          &           & 10.29          & 0.5306          & 0.10342          & 0.4737          \\
MasaCtrl &  & 23.36          &               & 0.1012           &           & 20.58          &           & 13.74          & 0.3671          & 0.05250          & 0.6645          \\
FPE      &  & 24.42          &                 & 0.1172          &           & \textbf{22.94}          &           & 11.66          & 0.4677          & 0.08009          & 0.5194          \\
InfEdit  &  & 21.98          &                 & 0.0504          &           & 22.71          &           & 15.34          & 0.2495          & 0.04359          & 0.7057          \\
 \cline{1-1} \cline{3-3} \cline{5-5} \cline{7-7} \cline{9-12} 
ALE     &  & \textbf{16.41} & & \textbf{0.0088} & \textbf{} & 22.62 & \textbf{} & \textbf{30.01} & \textbf{0.0405} & \textbf{0.00167} & \textbf{0.9049} \\ 
\bottomrule

\end{tabular}
}

\caption{
Quantitative evaluation of \textbf{editing one object} for ALE and baselines on \benchmark{}.
}

\label{table:object1}
\end{table*}
\begin{table*}[p]\centering
\resizebox{\textwidth}{!}{%
\begin{tabular}{cccccccccccccc}
\toprule
\multirow{2}{*}{\textbf{Method}} &
&
\multirow{2}{*}{\textbf{TELS ↓}}
 &
 &
 \multirow{2}{*}{\textbf{TILS ↓}} &
 &
\multirow{2}{*}{\textbf{\begin{tabular}[c]{@{}c@{}}Structure \\ Distance ↓\end{tabular}}}
&
&
\multirow{2}{*}{\textbf{\begin{tabular}[c]{@{}c@{}}Editing \\ Performance ↑\end{tabular}}}
 && \multicolumn{4}{c}{\textbf{\begin{tabular}[c]{@{}c@{}}Background  Preservation\end{tabular}}} \\
 && & & & & & & & & PSNR ↑& LPIPS ↓ & MSE ↓& SSIM ↑\\ \cline{1-1} \cline{3-3} \cline{5-5} \cline{7-7} \cline{9-9} \cline{11-14}


 P2P      &  & 20.87          &           & 17.52          &           & 0.1499           &           & 20.41          &           & 11.11          & 0.4506          & 0.08699          & 0.5560          \\
MasaCtrl &  & 19.58          &           & 16.90          &           & 0.0911           &           & 19.92          &           & 14.99          & 0.2886          & 0.04058          & 0.7357          \\
FPE      &  & 20.44          &           & 17.68          &           & 0.1141          &           & 21.72          &           & 12.81          & 0.3880          & 0.06439          & 0.6050          \\
InfEdit  &  & 19.16          &           & 16.86          &           & 0.0485          &           & 21.52          &           & 16.69          & 0.2026          & 0.03288          & 0.7719          \\
\cline{1-1} \cline{3-3} \cline{5-5} \cline{7-7} \cline{9-9} \cline{11-14} 
ALE     &  & \textbf{16.00} & \textbf{} & \textbf{15.42} & \textbf{} & \textbf{0.0165} & \textbf{} & \textbf{22.06} & \textbf{} & \textbf{30.06} & \textbf{0.0360} & \textbf{0.00146} & \textbf{0.9235} \\
\bottomrule

\end{tabular}
}

\caption{
Quantitative evaluation of \textbf{editing two objects} for ALE and baselines on \benchmark{}.
}

\label{table:object2}
\end{table*}
\begin{table*}[p]\centering

\resizebox{\textwidth}{!}{%
\begin{tabular}{cccccccccccccc}
\toprule
\multirow{2}{*}{\textbf{Method}} &
&
\multirow{2}{*}{\textbf{TELS ↓}}
 &
 &
 \multirow{2}{*}{\textbf{TILS ↓}} &
 &
\multirow{2}{*}{\textbf{\begin{tabular}[c]{@{}c@{}}Structure \\ Distance ↓\end{tabular}}}
&
&
\multirow{2}{*}{\textbf{\begin{tabular}[c]{@{}c@{}}Editing \\ Performance ↑\end{tabular}}}
 && \multicolumn{4}{c}{\textbf{\begin{tabular}[c]{@{}c@{}}Background  Preservation\end{tabular}}} \\
 && & & & & & & & & PSNR ↑& LPIPS ↓ & MSE ↓& SSIM ↑\\ \cline{1-1} \cline{3-3} \cline{5-5} \cline{7-7} \cline{9-9} \cline{11-14} 
P2P      &  & 18.80          &           & 17.00          &           & 0.1531          &           & 20.05          &           & 12.05          & 0.3674          & 0.07318          & 0.6469          \\
MasaCtrl &  & 17.60          &           & 16.58          &           & 0.0866          &           & 19.53          &           & 16.26          & 0.2231          & 0.03245          & 0.8037          \\
FPE      &  & 18.36          &           & 17.09          &           & 0.1180          &           & 20.99          &           & 14.00          & 0.3153          & 0.05247          & 0.6911          \\
InfEdit  &  & 17.62          &           & 16.51          &           & 0.0463          &           & 21.10          &           & 18.18          & 0.1580          & 0.02540          & 0.8350          \\
\cline{1-1} \cline{3-3} \cline{5-5} \cline{7-7} \cline{9-9} \cline{11-14} 
ALE     &  & \textbf{15.89} & \textbf{} & \textbf{15.36} & \textbf{} & \textbf{0.0246} & \textbf{} & \textbf{22.19} & \textbf{} & \textbf{30.01} & \textbf{0.0323} & \textbf{0.00154} & \textbf{0.9426} \\
\bottomrule
\end{tabular}
}

\caption{
Quantitative evaluation of \textbf{editing three objects} for ALE and baselines on \benchmark{}.
}
\label{table:object3}
\end{table*}
Tables~\ref{table:object1}, \ref{table:object2}, and \ref{table:object3} present the quantitative evaluation of our method and baselines on \benchmark{} across different numbers of editing objects.
For the baseline methods, TELS and TILS decrease as the number of edited objects increases, as editing more objects provides a more detailed description of the image, reducing ambiguity.
This trend highlights the baselines' dependence on long and detailed prompts.
However, their editing performance decreases with an increasing number of edited objects, revealing their limitations in handling complex edits.
In contrast, our method demonstrates robust performance across all object counts, consistently achieving the lowest leakage values, preserving structure and background, and maintaining competitive or superior editing performance.

\begin{figure*}[p]
    \centering
    \resizebox{0.8\textwidth}{!}{%
        \includegraphics[]{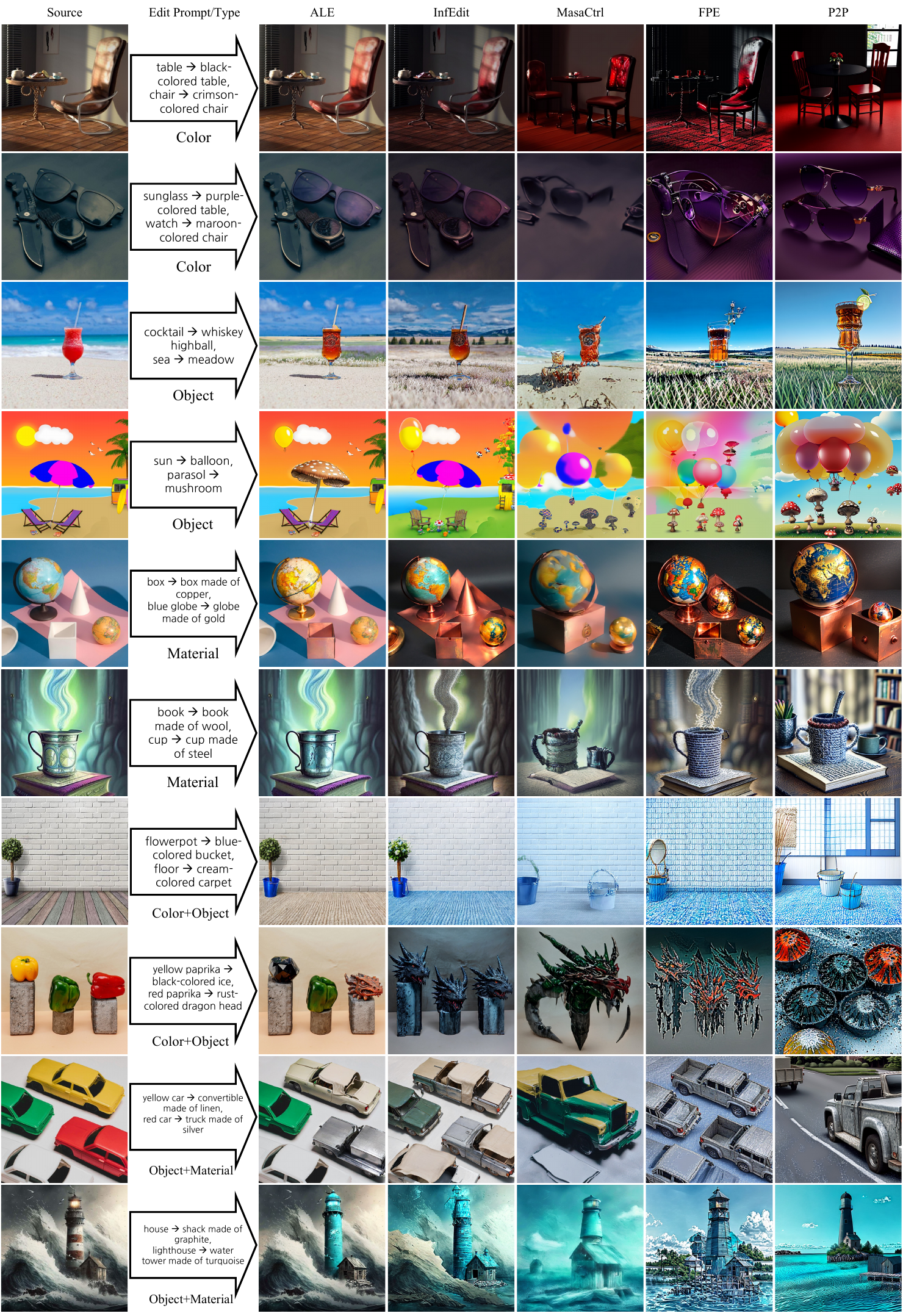}
    }
     \caption{
     Qualitative examples of editing results for each edit type on \benchmark{}. Two examples are provided for each edit type.
     The left side of $\xrightarrow{}$ represents the {source} prompt, and the right side represents the {target} prompt.
    }
    \label{fig:more_comparisons}
\end{figure*}

\paragraph{By edit type}
\begin{table*}[p]\centering

\resizebox{\textwidth}{!}{%
\begin{tabular}{cccccccccccccc}
\toprule
\multirow{2}{*}{\textbf{Method}} &
&
\multirow{2}{*}{\textbf{TELS ↓}}
 &
 &
 \multirow{2}{*}{\textbf{TILS ↓}} &
 &
\multirow{2}{*}{\textbf{\begin{tabular}[c]{@{}c@{}}Structure \\ Distance ↓\end{tabular}}}
&
&
\multirow{2}{*}{\textbf{\begin{tabular}[c]{@{}c@{}}Editing \\ Performance ↑\end{tabular}}}
 && \multicolumn{4}{c}{\textbf{\begin{tabular}[c]{@{}c@{}}Background  Preservation\end{tabular}}} \\
 && & & & & & & & & PSNR ↑& LPIPS ↓ & MSE ↓& SSIM ↑\\ \cline{1-1} \cline{3-3} \cline{5-5} \cline{7-7} \cline{9-9} \cline{11-14} 


P2P      &  & 23.20          &           & 18.02          &           & 0.1467           &           & 21.94          &           & 11.03          & 0.4529          & 0.0898           & 0.5753          \\
MasaCtrl &  & 21.70          &           & 17.35          &           & 0.0964           &           & 21.64          &           & 14.59          & 0.3113          & 0.04557          & 0.7283          \\
FPE      &  & 22.33          &           & 17.94          &           & 0.1065          &           & \textbf{23.23}         &           & 12.66          & 0.3926          & 0.06901          & 0.6266          \\
InfEdit  &  & 19.68          &           & 17.31          &           & 0.0343          &           & 23.16          &           & 18.54          & 0.1401          & 0.02695          & 0.8347          \\
\cline{1-1} \cline{3-3} \cline{5-5} \cline{7-7} \cline{9-9} \cline{11-14} 
ALE     &  & \textbf{17.63} & \textbf{} & \textbf{16.21} & \textbf{} & \textbf{0.0089} & \textbf{} & 23.12 & \textbf{} & \textbf{32.97} & \textbf{0.0288} & \textbf{0.00079} & \textbf{0.9309} \\
\bottomrule

\end{tabular}
}

\caption{
Quantitative evaluation of the \textbf{color change} edit type for ALE and baselines on \benchmark{}.
}
\label{table:color}
\end{table*}
\begin{table*}[p]\centering

\resizebox{\textwidth}{!}{%
\begin{tabular}{cccccccccccccc}
\toprule
\multirow{2}{*}{\textbf{Method}} &
&
\multirow{2}{*}{\textbf{TELS ↓}}
 &
 &
 \multirow{2}{*}{\textbf{TILS ↓}} &
 &
\multirow{2}{*}{\textbf{\begin{tabular}[c]{@{}c@{}}Structure \\ Distance ↓\end{tabular}}}
&
&
\multirow{2}{*}{\textbf{\begin{tabular}[c]{@{}c@{}}Editing \\ Performance ↑\end{tabular}}}
 && \multicolumn{4}{c}{\textbf{\begin{tabular}[c]{@{}c@{}}Background  Preservation\end{tabular}}} \\
 && & & & & & & & & PSNR ↑& LPIPS ↓ & MSE ↓& SSIM ↑\\ \cline{1-1} \cline{3-3} \cline{5-5} \cline{7-7} \cline{9-9} \cline{11-14}


P2P      &  & 20.65          &           & 17.78          &           & 0.1535          &           & 20.38          &           & 11.23          & 0.4457          & 0.08750           & 0.5485          \\
MasaCtrl &  & 19.52          &           & 17.33          &           & 0.0901          &           & 19.72          &           & 15.48          & 0.2744          & 0.03801          & 0.7417          \\
FPE      &  & 19.76          &           & 17.58          &           & 0.1221          &           & 21.30          &           & 13.12          & 0.3732          & 0.06236          & 0.6103          \\
InfEdit  &  & 18.67          &           & 17.12          &           & 0.0504          &           & 21.10          &           & 16.59          & 0.2114          & 0.03381          & 0.7607          \\
\cline{1-1} \cline{3-3} \cline{5-5} \cline{7-7} \cline{9-9} \cline{11-14} 
ALE     &  & \textbf{15.86} & \textbf{} & \textbf{16.25} & \textbf{} & \textbf{0.0197} & \textbf{} & \textbf{21.82} & \textbf{} & \textbf{29.03} & \textbf{0.0386} & \textbf{0.00182} & \textbf{0.9218} \\
\bottomrule

\end{tabular}
}

\caption{
Quantitative evaluation of the \textbf{object change} edit type for ALE and baselines on \benchmark{}.
}

\label{table:object}
\end{table*}
\begin{table*}[p]\centering

\resizebox{\textwidth}{!}{%
\begin{tabular}{cccccccccccccc}
\toprule
\multirow{2}{*}{\textbf{Method}} &
&
\multirow{2}{*}{\textbf{TELS ↓}}
 &
 &
 \multirow{2}{*}{\textbf{TILS ↓}} &
 &
\multirow{2}{*}{\textbf{\begin{tabular}[c]{@{}c@{}}Structure \\ Distance ↓\end{tabular}}}
&
&
\multirow{2}{*}{\textbf{\begin{tabular}[c]{@{}c@{}}Editing \\ Performance ↑\end{tabular}}}
 && \multicolumn{4}{c}{\textbf{\begin{tabular}[c]{@{}c@{}}Background  Preservation\end{tabular}}} \\
 && & & & & & & & & PSNR ↑& LPIPS ↓ & MSE ↓& SSIM ↑\\ \cline{1-1} \cline{3-3} \cline{5-5} \cline{7-7} \cline{9-9} \cline{11-14} 

P2P      &  & 21.40          &           & 17.07          &           & 0.1549          &           & 20.75          &           & 11.07          & 0.4417          & 0.08674          & 0.5429          \\
MasaCtrl &  & 20.82          &           & 17.00          &           & 0.0866          &           & 20.93          &           & 15.39          & 0.2755          & 0.03763          & 0.7418          \\
FPE      &  & 21.91          &           & 17.68          &           & 0.1151          &           & 22.64          &           & 13.14          & 0.3781          & 0.05908          & 0.5943          \\
InfEdit  &  & 20.33          &           & 16.87          &           & 0.0387          &           & 22.59          &           & 17.71          & 0.1753          & 0.02559          & 0.7862          \\
\cline{1-1} \cline{3-3} \cline{5-5} \cline{7-7} \cline{9-9} \cline{11-14} 
ALE     &  & \textbf{17.15} & \textbf{} & \textbf{15.96} & \textbf{} & \textbf{0.0118} & \textbf{} & \textbf{22.94} & \textbf{} & \textbf{30.63} & \textbf{0.0339} & \textbf{0.00120} & \textbf{0.9248} \\
\bottomrule

\end{tabular}
}

\caption{
Quantitative evaluation of the \textbf{material change} edit type for ALE and baselines on \benchmark{}.
}

\label{table:material}
\end{table*}
\begin{table*}[p]\centering
\resizebox{\textwidth}{!}{%
\begin{tabular}{cccccccccccccc}
\toprule
\multirow{2}{*}{\textbf{Method}} &
&
\multirow{2}{*}{\textbf{TELS ↓}}
 &
 &
 \multirow{2}{*}{\textbf{TILS ↓}} &
 &
\multirow{2}{*}{\textbf{\begin{tabular}[c]{@{}c@{}}Structure \\ Distance ↓\end{tabular}}}
&
&
\multirow{2}{*}{\textbf{\begin{tabular}[c]{@{}c@{}}Editing \\ Performance ↑\end{tabular}}}
 && \multicolumn{4}{c}{\textbf{\begin{tabular}[c]{@{}c@{}}Background  Preservation\end{tabular}}} \\
 && & & & & & & & & PSNR ↑& LPIPS ↓ & MSE ↓& SSIM ↑\\ \cline{1-1} \cline{3-3} \cline{5-5} \cline{7-7} \cline{9-9} \cline{11-14} 

P2P      &  & 21.62          &           & 17.10          &           & 0.1483          &           & 20.71          &           & 11.12          & 0.4586          & 0.0896           & 0.5786          \\
MasaCtrl &  & 19.59          &           & 16.15          &           & 0.0986          &           & 19.18          &           & 14.43          & 0.3142          & 0.04634          & 0.7281          \\
FPE      &  & 20.82          &           & 17.11          &           & 0.1159          &           & 21.40          &           & 12.42          & 0.4079          & 0.07175          & 0.6081          \\
InfEdit  &  & 19.92          &           & 16.11          &           & 0.0619          &           & 21.23          &           & 15.04          & 0.2543          & 0.04503          & 0.7333          \\
\cline{1-1} \cline{3-3} \cline{5-5} \cline{7-7} \cline{9-9} \cline{11-14} 
ALE     &  & \textbf{15.30} & \textbf{} & \textbf{14.01} & \textbf{} & \textbf{0.0231} & \textbf{} & \textbf{22.15} & \textbf{} & \textbf{28.60} & \textbf{0.0407} & \textbf{0.00205} & \textbf{0.9206} \\
\bottomrule

\end{tabular}
}

\caption{
Quantitative evaluation of the \textbf{color and object change} edit type for ALE and baselines on \benchmark{}.
}

\label{table:color_object}
\end{table*}
\begin{table*}[p]\centering

\resizebox{\textwidth}{!}{%
\begin{tabular}{cccccccccccccc}
\toprule
\multirow{2}{*}{\textbf{Method}} &
&
\multirow{2}{*}{\textbf{TELS ↓}}
 &
 &
 \multirow{2}{*}{\textbf{TILS ↓}} &
 &
\multirow{2}{*}{\textbf{\begin{tabular}[c]{@{}c@{}}Structure \\ Distance ↓\end{tabular}}}
&
&
\multirow{2}{*}{\textbf{\begin{tabular}[c]{@{}c@{}}Editing \\ Performance ↑\end{tabular}}}
 && \multicolumn{4}{c}{\textbf{\begin{tabular}[c]{@{}c@{}}Background  Preservation\end{tabular}}} \\
 && & & & & & & & & PSNR ↑& LPIPS ↓ & MSE ↓& SSIM ↑\\ \cline{1-1} \cline{3-3} \cline{5-5} \cline{7-7} \cline{9-9} \cline{11-14} 

P2P      &  & 20.76          &           & 16.31          &           & 0.1538          &           & 19.54          &           & 11.29          & 0.4488          & 0.0857          & 0.5490          \\
MasaCtrl &  & 19.28          &           & 15.86          &           & 0.0929          &           & 18.57          &           & 15.07          & 0.2893          & 0.04166         & 0.7332          \\
FPE      &  & 20.55          &           & 16.62          &           & 0.1224          &           & 20.86          &           & 12.77          & 0.3998          & 0.06604         & 0.5865          \\
InfEdit  &  & 19.34          &           & 16.03          &           & 0.0567          &           & 20.80          &           & 15.81          & 0.2356          & 0.03842         & 0.7394          \\
\cline{1-1} \cline{3-3} \cline{5-5} \cline{7-7} \cline{9-9} \cline{11-14} 
ALE     &  & \textbf{14.55} & \textbf{} & \textbf{14.51} & \textbf{} & \textbf{0.0196} & \textbf{} & \textbf{21.42} & \textbf{} & \textbf{28.88} & \textbf{0.0393} & \textbf{0.0019} & \textbf{0.9201} \\
\bottomrule

\end{tabular}
}

\caption{
Quantitative evaluation of the \textbf{object and material change} edit type for ALE and baselines on \benchmark{}.
}

\label{table:object_material}
\end{table*}
We compare our methods with baselines across different edit types in Tables~\ref{table:color}, \ref{table:object}, \ref{table:material}, \ref{table:color_object}, and \ref{table:object_material}.
Across all edit types, our method consistently outperforms baselines by achieving lower leakage, better structural and background preservation, and strong editing performance.
We provide more qualitative examples on \benchmark{} for each edit type in two objects editing in Figure~\ref{fig:more_comparisons}.


\section{Ablation Study Results} 
\label{appendix:ablation}

\paragraph{Ablation on EOS embedding methods}
\begin{figure*}[ht]
    \centering
    \includegraphics[width=0.95\textwidth]{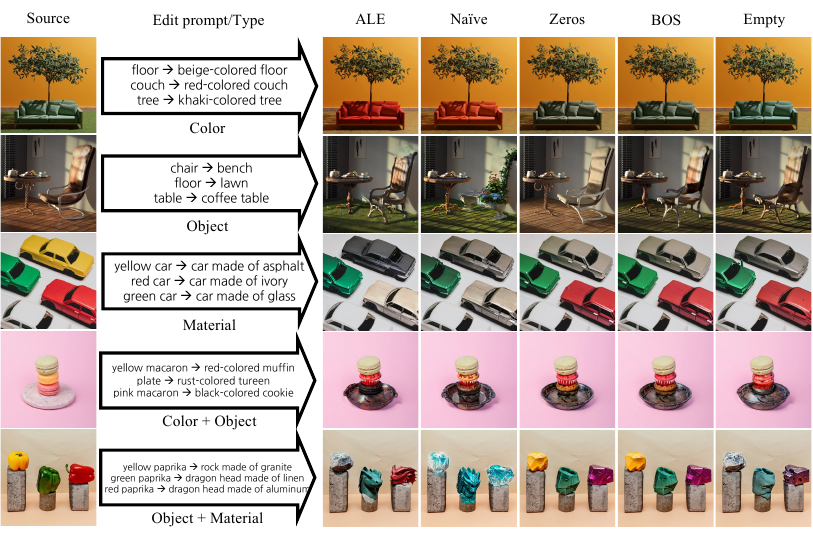}
    \caption{
    Qualitative examples from the EOS ablation study.
    While our method produces convincing results, other methods fail to generate the target object or exhibit attribute leakage.
    For instance, using the naive EOS to edit an object generates plants in place of the chair.
    This occurs due to attribute leakage from the word {lawn} to {bench}, resulting in chair-shaped flowers.
    }
    \label{fig:EOS ablation}
\end{figure*}
\begin{table*}[p]
\centering

\resizebox{\textwidth}{!}{%
\begin{tabular}{cccccccccccccc}
\toprule
\multirow{2}{*}{\textbf{Method}} &
&
\multirow{2}{*}{\textbf{TELS ↓}}
 &
 &
 \multirow{2}{*}{\textbf{TILS ↓}} &
 &
\multirow{2}{*}{\textbf{\begin{tabular}[c]{@{}c@{}}Structure \\ Distance ↓\end{tabular}}}
&
&
\multirow{2}{*}{\textbf{\begin{tabular}[c]{@{}c@{}}Editing \\ Performance ↑\end{tabular}}}
 && \multicolumn{4}{c}{\textbf{\begin{tabular}[c]{@{}c@{}}Background  Preservation\end{tabular}}} \\
 && & & & & & & & & PSNR ↑& LPIPS ↓ & MSE ↓& SSIM ↑\\ \cline{1-1} \cline{3-3} \cline{5-5} \cline{7-7} \cline{9-9} \cline{11-14} 

Naive&& 16.02 && 15.81&& 0.0156 && 21.86 && 30.14& 0.0359& 0.0014& 0.9232\\ 
Zeros&& \textbf{15.74}& & \textbf{15.23}& & \textbf{0.0107} & & 20.78& & \textbf{31.22} & \textbf{0.0327} & \textbf{0.0011} & \textbf{0.9254} \\
BOS&& 15.76 & & 15.27 & & 0.0115& & 20.87 & & 31.09& 0.0334& 0.0011& 0.9241\\
Empty String && 15.86&& 15.33&& 0.0139 && 21.25 && 30.61 & 0.0342 & 0.0013 & 0.9248 \\ \cline{1-1} \cline{3-3} \cline{5-5} \cline{7-7} \cline{9-9} \cline{11-14} 
ALE && 16.03 & & 15.28 & & 0.0167& & \textbf{22.20} & & 30.04& 0.0361& 0.0014& 0.9228   \\ 
\bottomrule

\end{tabular}
}

\caption{
    Ablation study on different strategies for handling EOS embeddings in the prompt.
    While ALE shows slightly higher leakages compared to others, it achieves the best editing performance. 
    All experiments were conducted with both RGB-CAM and BB applied.
}
\label{table:ablation_eos}
\end{table*}
To evaluate the effect of EOS embeddings, we studied several methods of modifying EOS embeddings:
(1) Naive: No modification, using the original EOS embeddings;
(2) Zeros: Replacing EOS embeddings with zero-valued vectors;
(3) BOS: Substituting EOS embeddings with BOS (beginning-of-sequence) embeddings;
(4) Empty String: Using EOS embeddings derived from an empty string.
In Figure~\ref{fig:EOS_modification}, our method demonstrates robust results across various scenarios, while the other methods often produce images that fail to follow the edit prompt or exhibit attribute leakage.
A detailed quantitative comparison is provided in Table~\ref{table:ablation_eos}.

Another EOS modification method is proposed in \cite{hu2024token}, named End Token Substitution (ETS).
ETS substitutes an embedding of EOS in a full prompt into an embedding of EOS in a rephrased prompt, which deletes all attribute expressions, e.g. ``a yellow cat and a white dog'' into ``a cat and a dog''.
\begin{figure}[!ht]
    \includegraphics[width=0.48\textwidth]{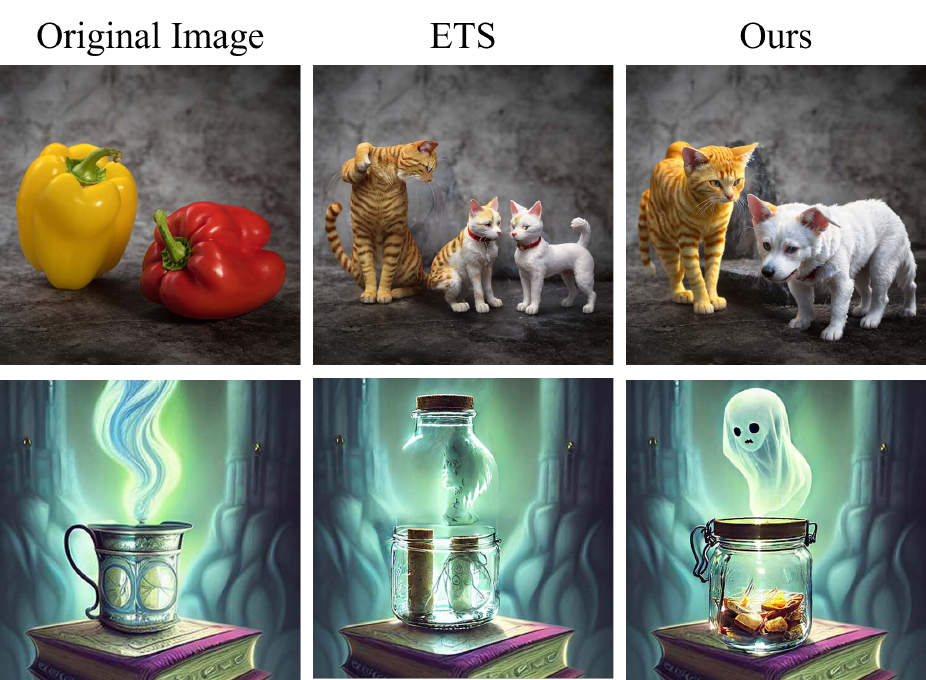}
    \vspace{-0.5cm}
    \caption{
    ETS, which is proposed in \cite{hu2024token}, fails to generate intended results. 
    Edit prompt: (up) yellow paprika $\xrightarrow{}$ {yellow cat}, {red paprika} $\xrightarrow{}$ {white dog}. 
    (down) {cup} $\xrightarrow{}$ {jar}, {steam} $\xrightarrow{}$ {ghost}.
    }
    \label{fig:ablation_ETS}
\end{figure}
In Figure~\ref{fig:ablation_ETS}, TI leakage in ETS is observed, e.g. cats are generated instead of a cat and a dog, and jar is generated in a region where ghost should be.
RGB-CAM is applied for both methods, therefore the cross-attention mappings are aligned with the prompt.

Our method consistently achieves the best editing performance while maintaining competitive structure and background preservation metrics.
In contrast, the other methods reveal a trade-off between reducing leakage and maintaining high editing performance, highlighting the effectiveness of our approach in balancing these objectives.

\paragraph{Ablation on RGB-CAM and BB}
\begin{table*}[t]\centering

\resizebox{\textwidth}{!}{%
\begin{tabular}{cccccccccccccc}
\toprule
\multirow{2}{*}{\textbf{Method}} &
&
\multirow{2}{*}{\textbf{TELS ↓}}
 &
 &
 \multirow{2}{*}{\textbf{TILS ↓}} &
 &
\multirow{2}{*}{\textbf{\begin{tabular}[c]{@{}c@{}}Structure \\ Distance ↓\end{tabular}}}
&
&
\multirow{2}{*}{\textbf{\begin{tabular}[c]{@{}c@{}}Editing \\ Performance ↑\end{tabular}}}
 && \multicolumn{4}{c}{\textbf{\begin{tabular}[c]{@{}c@{}}Background  Preservation\end{tabular}}} \\
 && & & & & & & & & PSNR ↑& LPIPS ↓ & MSE ↓& SSIM ↑\\ \cline{1-1} \cline{3-3} \cline{5-5} \cline{7-7} \cline{9-9} \cline{11-14} 
 
ORE&& 20.05& & 16.87& & 0.0521 & & 21.81& & 16.16 & 0.2182 & 0.0380 & 0.7591 \\
ORE+RGB&& 18.99 & & 15.46 & & 0.0436& & \textbf{22.42} & & 17.48& 0.1805& 0.0291& 0.7887\\
ORE+BB && 16.12 &  & 16.58&  & \textbf{0.0164} &  & 21.56 &  & 29.88& 0.0368& 0.0015& 0.9219\\ \cline{1-1} \cline{3-3} \cline{5-5} \cline{7-7} \cline{9-9} \cline{11-14} 
ALE && \textbf{16.03}&  & \textbf{15.28}&  & 0.0167 &  & 22.20 &  & \textbf{30.04} & \textbf{0.0361} & \textbf{0.0014} & \textbf{0.9228} \\ \bottomrule

\end{tabular}
}

\caption{
    Ablation study comparing the components of our method: object-restricted embeddings (ORE), region-guided blending cross-attention masking (RGB), and background blending (BB). 
    RGB markedly reduces TILS, whereas BB substantially lowers TELS. 
    When ORE is used \emph{without} RGB, it relies solely on the base embedding $E'_{\text{base}}$ (i.e., the {ORE} and {ORE\,+\,BB} cases). 
    Integrating all three components ({ALE}) yields the best overall performance across nearly every metric, underscoring their complementary strengths.
}
\label{table:ablation_cam_bb}
\end{table*}
The results in Table~\ref{table:ablation_cam_bb} demonstrate the complementary strengths of RGB-CAM and BB in our method.
While RGB-CAM effectively reduces TI leakage by confining edits to the targeted objects, its impact on TE leakage and background preservation is limited.
Conversely, BB significantly lowers TE leakage by preserving non-target regions, improving background quality but slightly reducing editing performance.
Combining all components (Ours) achieves the best overall balance, minimizing leakage while preserving structure and background, and maintaining strong editing performance, highlighting the synergy of these components.

\begin{table*}[!ht]\centering

\resizebox{0.8\textwidth}{!}{%
\begin{tabular}{cccccccccccccc}
\toprule
\multirow{2}{*}{\textbf{Method}} &
&
\multirow{2}{*}{\textbf{TELS ↓}}
 &
 &
\multirow{2}{*}{\textbf{\begin{tabular}[c]{@{}c@{}}Structure \\ Distance ↓\end{tabular}}}
&
&
\multirow{2}{*}{\textbf{\begin{tabular}[c]{@{}c@{}}Editing \\ Performance ↑\end{tabular}}}
 && \multicolumn{4}{c}{\textbf{\begin{tabular}[c]{@{}c@{}}Background  Preservation\end{tabular}}} \\
 && & & & & & & PSNR ↑& LPIPS ↓ & MSE ↓& SSIM ↑\\ \cline{1-1} \cline{3-3} \cline{5-5} \cline{7-7} \cline{9-12} 
 
P2P&& 26.20 & & 0.1571& & 23.74& & 11.11& 0.4270& 0.0919& 0.4600\\
MasaCtrl && 24.48 & & 0.0856& & 22.16& & 15.81& 0.2540& 0.0334& 0.6803\\
FPE&& 25.64 & & 0.1265& & \textbf{23.89} & & 13.35& 0.3499& 0.0581& 0.5346\\
InfEdit&& 24.51 & & 0.0446& & 22.92& & 19.41& 0.1519& 0.0168& 0.7581\\ \cline{1-1} \cline{3-3} \cline{5-5} \cline{7-7} \cline{9-12} 
ALE && \textbf{22.94}&& \textbf{0.0238} && 22.87&& \textbf{28.77} & \textbf{0.0580} & \textbf{0.0046} & \textbf{0.8865} \\ 
\bottomrule

\end{tabular}
}

\caption{
Evaluation results on PIE-Bench for compatible edit types (object change, content change, color change, and material change).
Our method achieves the lowest TELS and demonstrates the best structure and background preservation while maintaining competitive editing performance.
}
\label{table:pie_results}
\end{table*}

\paragraph{Evaluation on PIE-Bench}
We also evaluated our method on the existing PIE-Bench \cite{ju2023direct} in addition to \benchmark{}.
Since our method does not support all edit types in PIE-Bench, we conducted experiments on the four edit types that are compatible: \textit{object change}, \textit{content change}, \textit{color change}, and \textit{material change}.

PIE-Bench only considers scenarios with a single object editing, so we excluded the TI Leakage metric.
When running our method, we used the blend word provided by PIE-Bench as the SAM prompt for mask generation.
In cases where mask segmentation failed, we edited the image without cross-attention masking and background blending.

\begin{figure*}[p]
    \centering
    \resizebox{\textwidth}{!}{%
        \includegraphics[]{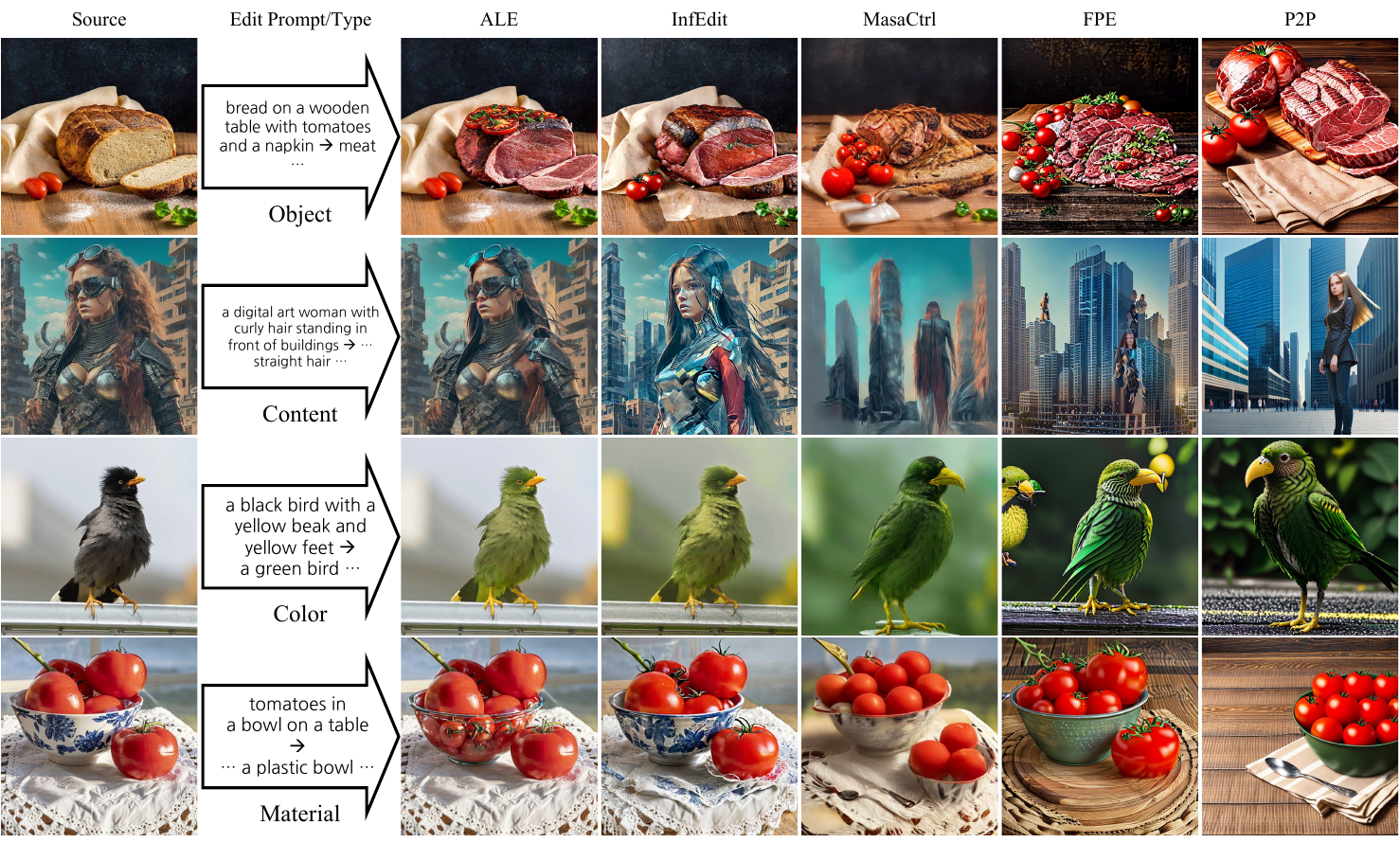}
    }
     \caption{
     Qualitative examples of editing results for the four compatible edit types on PIE-Bench: object change, content change, color change, and material change.
     In edit prompt column, the left side of the arrow $\xrightarrow[]{}$ represents the source prompt, and the right side represents the target prompt, with unchanged parts omitted as ``...'' for brevity.
     Baseline methods exhibit attribute leakage or fail to preserve the source image structure, while our method achieves more precise edits with minimal leakage.
    }
    \label{fig:pie_examples}
\end{figure*}

In Table~\ref{table:pie_results}, the results show that our method demonstrated the lowest attribute leakage and high editing performance among all methods, even on PIE-Bench.
These findings further validate the robustness and versatility of our approach across different benchmarks.
We also provide qualitative examples for each edit type from the PIE-Bench experiments in Figure~\ref{fig:pie_examples}.

\paragraph{Ablation on self-attention injection schedule}
\begin{figure*}[t]
    \centering
    \includegraphics[width=\textwidth]{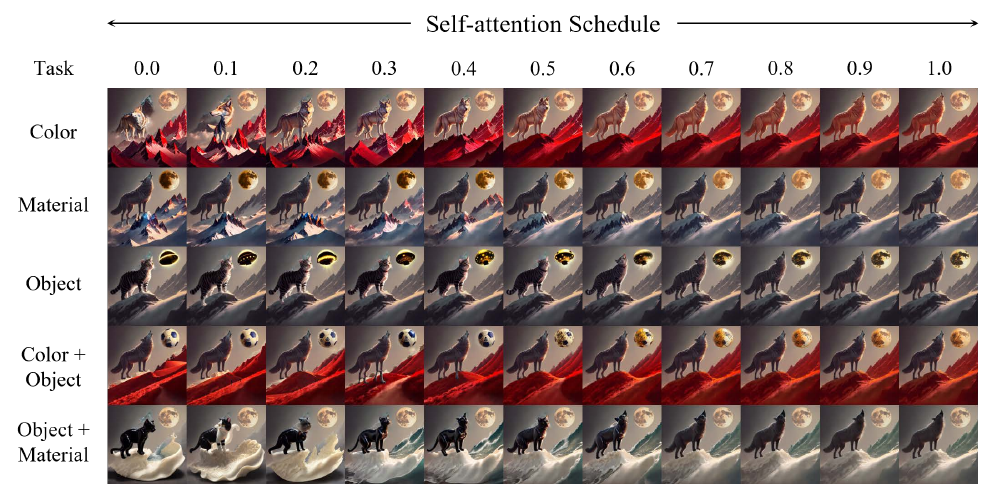}
    \caption{
    Ablation study on self-attention injection schedule.
    A schedule value specifies the fraction of early denoising steps during which self-attention maps from the source image are injected (e.g., 0.3 $\xrightarrow{}$ first 30 $\%$ of steps). 
    Larger values preserve more of the source structure and content, whereas smaller values grant greater freedom to satisfy the edit. 
    The optimal schedule therefore varies by edit type. 
    Prompts for each editing type are:
    (1) color: {wolf} $\xrightarrow{}$ {cream-colored} wolf, {mountain} $\xrightarrow{}$ {crimson-colored} mountain, 
    (2) material: {mountain} $\xrightarrow{}$ mountain {made of crystal}, {moon} $\xrightarrow{}$ moon {made of gold}, 
    (3) object: {wolf} $\xrightarrow{}$ {cat}, {moon} $\xrightarrow{}$  {UFO}, 
    (4) color + object: {moon} $\xrightarrow{}$ {navy-colored soccer ball}, {mountain} $\xrightarrow{}$ {crimson-colored hill}, 
    (5) object + material: {cat} $\xrightarrow{}$ wolf {made of rubber}, {mountain} $\xrightarrow{}$ {wave made of ivory}.
    }
    \label{fig:self-attention ablation}
\end{figure*}
The degree to which the structure of a source image needs to be preserved varies depending on the edit type.
For edits like color changes, maintaining the original structure is crucial, while object changes may require more deviation from the source. 
Figure~\ref{fig:ablation} shows the effect of the self-attention schedule across various scenarios.
Adjusting the schedule from 0.0 to 1.0 shows that higher values preserve more structure, while lower values allow greater flexibility.
Thus, selecting the appropriate self-attention schedule depends on the specific goals of the task. 
The hyperparameters we used were chosen based on these experimental findings.

\section{Limitations}
\label{appendix:limitations}

\paragraph{\benchmark{}}
While our benchmark provides a robust framework for evaluating attribute leakage in image editing, it has certain limitations.
First, the range of editing tasks is currently limited to basic and mixed edits such as color, object, and material changes.
More complex editing types, such as style transfer or pose modifications, are not covered in ALE-Bench.
However, defining attribute leakage in edits like style transfer is inherently ambiguous, as such edits often involve holistic changes to the image, making it unclear which regions should remain unaffected.
Addressing these challenges would require redefining attribute leakage for these contexts and designing new evaluation metrics tailored to these specific tasks.
Second, the dataset size (20 images) may limit the evaluation of models trained on larger or more diverse datasets.
Future updates of \benchmark{} could expand its scope by incorporating additional images, and more diverse editing types to overcome these limitations.

\begin{figure}[ht!]
    \centering
    \begin{subfigure}[b]{0.31\linewidth}
        \centering
        \includegraphics[width=\linewidth]{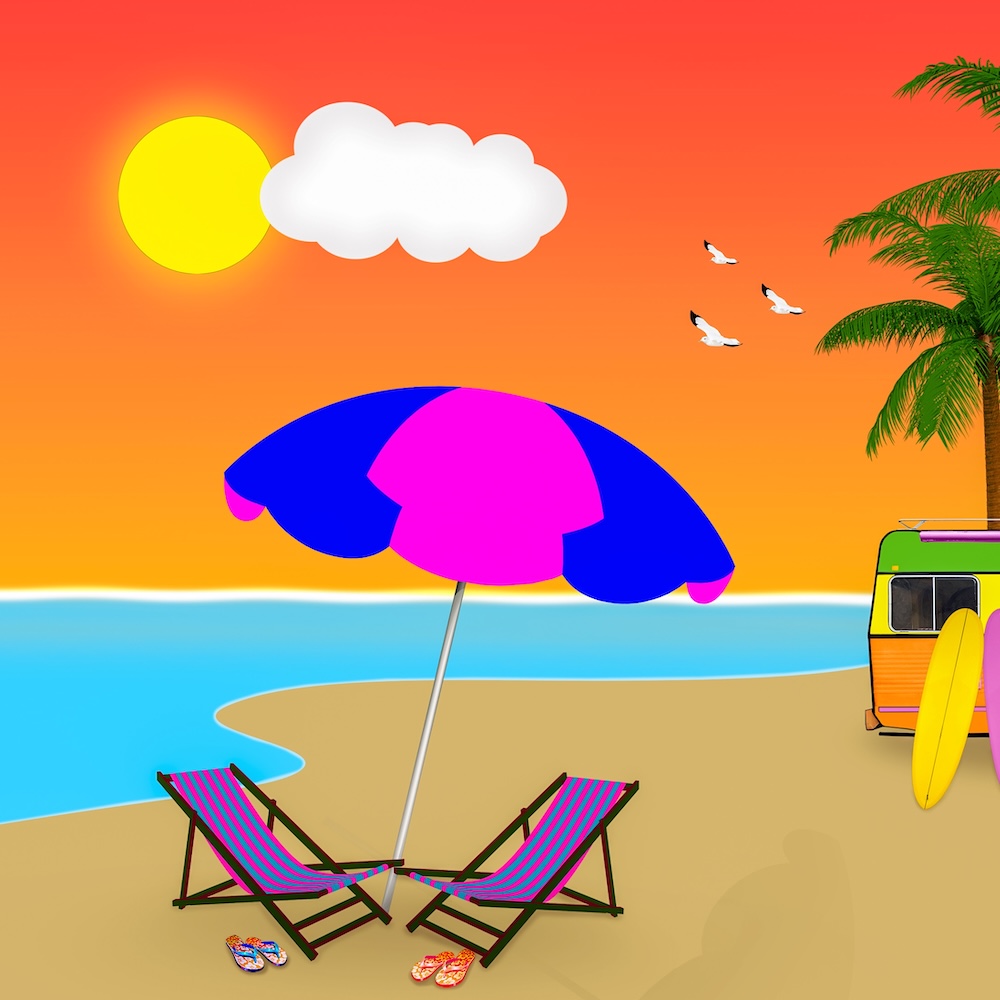}
        \caption{Source image}
        \label{fig:model_fail_source}
    \end{subfigure}
    \hfill
    \begin{subfigure}[b]{0.31\linewidth}
        \centering
        \includegraphics[width=\linewidth]{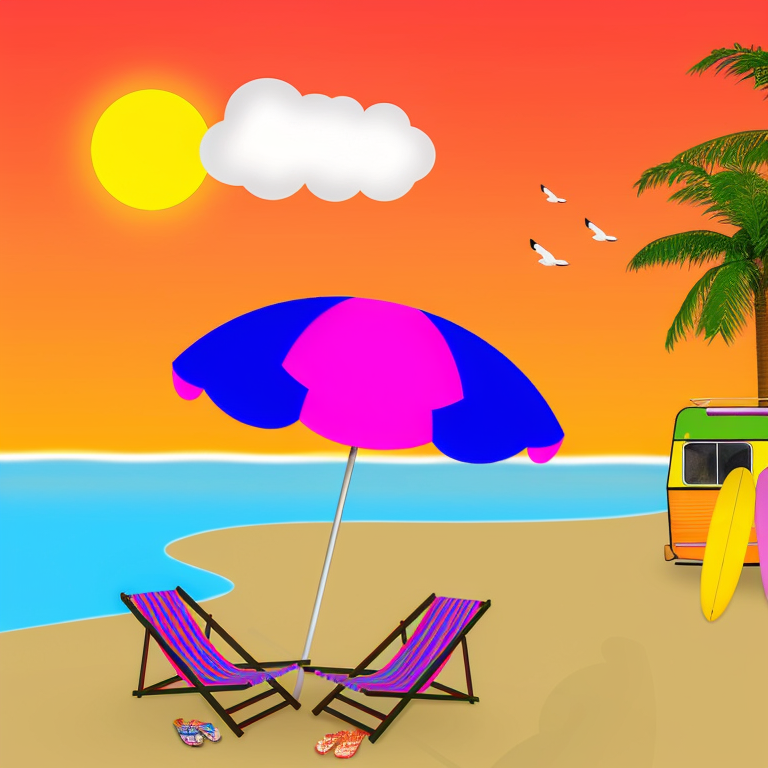}
        \caption{Editing result}
        \label{fig:model_fail_result}
    \end{subfigure}
    \hfill
    \begin{subfigure}[b]{0.31\linewidth}
        \centering
        \includegraphics[width=\linewidth]{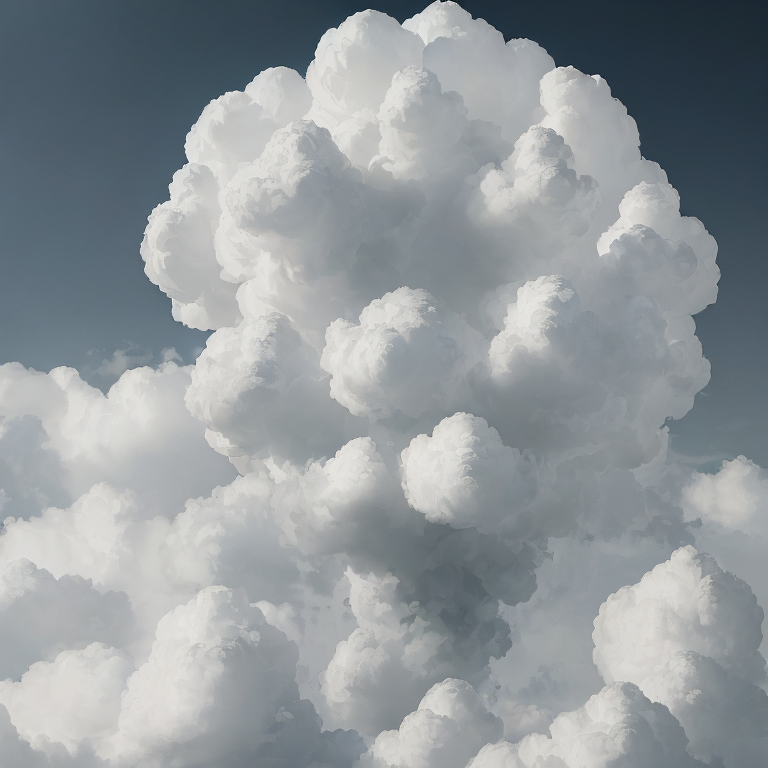}
        \caption{Generation result}
        \label{fig:model_fail_gen}
    \end{subfigure}
    \vspace{-0.3cm}
    \caption{Failure case due to the base model's inability.
    Editing prompt: {cloud} $\xrightarrow[]{}$ cloud made of {chrome}.
    Figure~\ref{fig:model_fail_gen}  illustrates the generation result when given the prompt ``cloud made of chrome''.
    }
    \label{figure:failure_model}
\end{figure}

\begin{figure}[ht]
    \centering
    \begin{subfigure}[b]{0.31\linewidth}
        \centering
        \includegraphics[width=\linewidth]{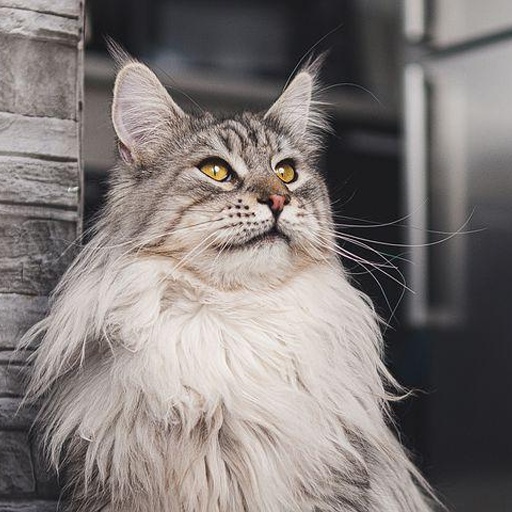}
        \caption{Source image}
        \label{fig:sam_source}
    \end{subfigure}
    \hfill
    \begin{subfigure}[b]{0.31\linewidth}
        \centering
        \includegraphics[width=\linewidth]{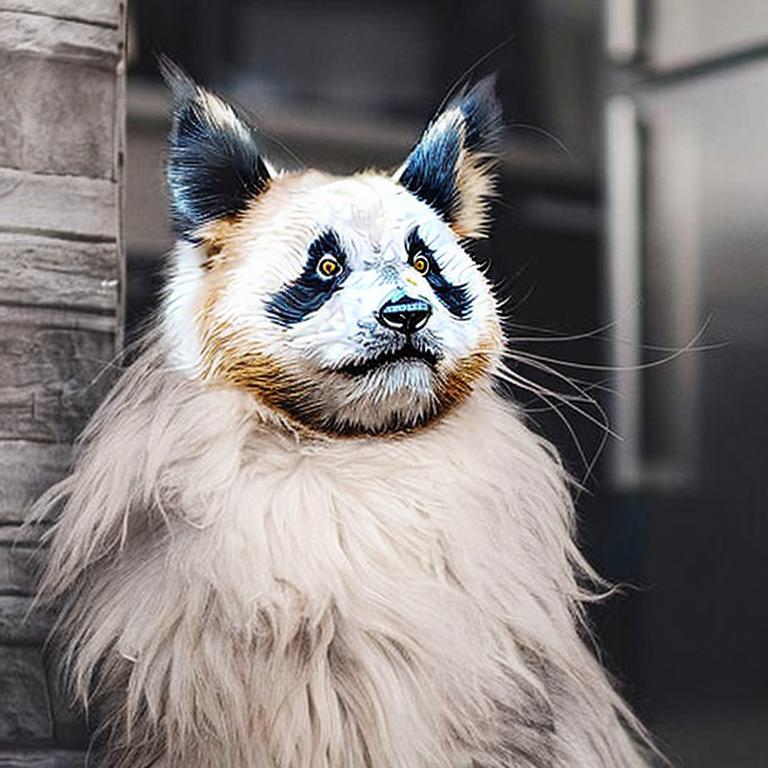}
        \caption{Editing result}
        \label{fig:sam_result}
    \end{subfigure}
    \hfill
    \begin{subfigure}[b]{0.31\linewidth}
        \centering
        \includegraphics[width=\linewidth]{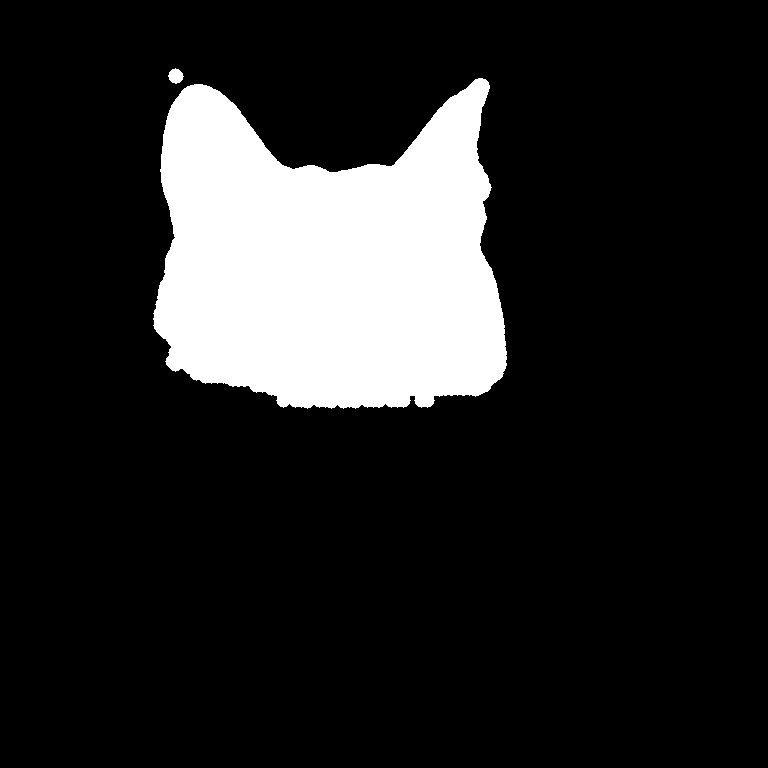}
        \caption{Segmentation mask}
        \label{fig:sam_mask}
    \end{subfigure}
    \vspace{-0.3cm}
    \caption{
        Failure case due to SAM segmentation fail.
        Editing prompt: $\ldots$ {cat} $\ldots$ $\xrightarrow[]{}$ $\ldots$ {panda} $\ldots$.
    Figure~\ref{fig:sam_mask}  shows the unsuccessful segmentation of SAM.
    }
    \label{figure:failure_sam}
\end{figure}

\paragraph{Failure cases}
Our framework leverages two backbone models, a pre-trained diffusion model and a segmentation model, Grounded-SAM.
Consequently, it may fail when the task exceeds the capabilities of these backbone models.
For instance, overly rare or complex prompts that the pre-trained diffusion model cannot handle (Figure~\ref{figure:failure_model}),
objects that are difficult for the segmentation model to recognize,
or incomplete segmentation masks generated by the model (Figure~\ref{figure:failure_sam}) can lead to unsatisfactory results.
However, since our method operates in parallel with advancements in these backbone models, we anticipate that such failure cases will decrease as these models continue to improve.



\end{document}